\documentclass{article} % For LaTeX2e
\usepackage{iclr2027_conference,times}

\usepackage{amsmath,amsfonts,bm}

\def\eqref#1{equation~\ref{#1}}
\def\1{\bm{1}}

\DeclareMathAlphabet{\mathsfit}{\encodingdefault}{\sfdefault}{m}{sl}
\SetMathAlphabet{\mathsfit}{bold}{\encodingdefault}{\sfdefault}{bx}{n}

\usepackage{hyperref}
\usepackage{url}

\usepackage{graphicx}
\usepackage{booktabs}       % professional-quality tables
\usepackage{amsfonts}       % blackboard math symbols
\usepackage{nicefrac}       % compact symbols for 1/2, etc.
\usepackage{xcolor}         % colors

\usepackage{amsmath}
\usepackage{amssymb}
\usepackage{graphicx}
\usepackage{multirow}
\usepackage{bm}
\usepackage{array}
\usepackage{makecell}
\usepackage{adjustbox}
\usepackage[table]{xcolor}
\usepackage{pifont}
\usepackage{algorithm}
\usepackage{algpseudocode}
\usepackage{enumitem}

\usepackage{cuted}
\usepackage{wrapfig}
\usepackage{caption}

\title{GeoSET: Generalist Foundation Model for SAR-to-EO Image Translation}

\author{
Jeonghyeok Do \quad Munchurl Kim \thanks{Corresponding author.}\\
Korea Advanced Institute of Science and Technology (KAIST)\\
\texttt{\{ehwjdgur0913,mkimee\}@kaist.ac.kr}\\[4pt]
{\small Project Page: \url{https://kaist-viclab.github.io/GeoSET_site/}}
}

\iclrfinalcopy
\begin{document}

\maketitle

\begin{center}
    \vspace{-0.8cm}
    \includegraphics[width=0.9\linewidth]{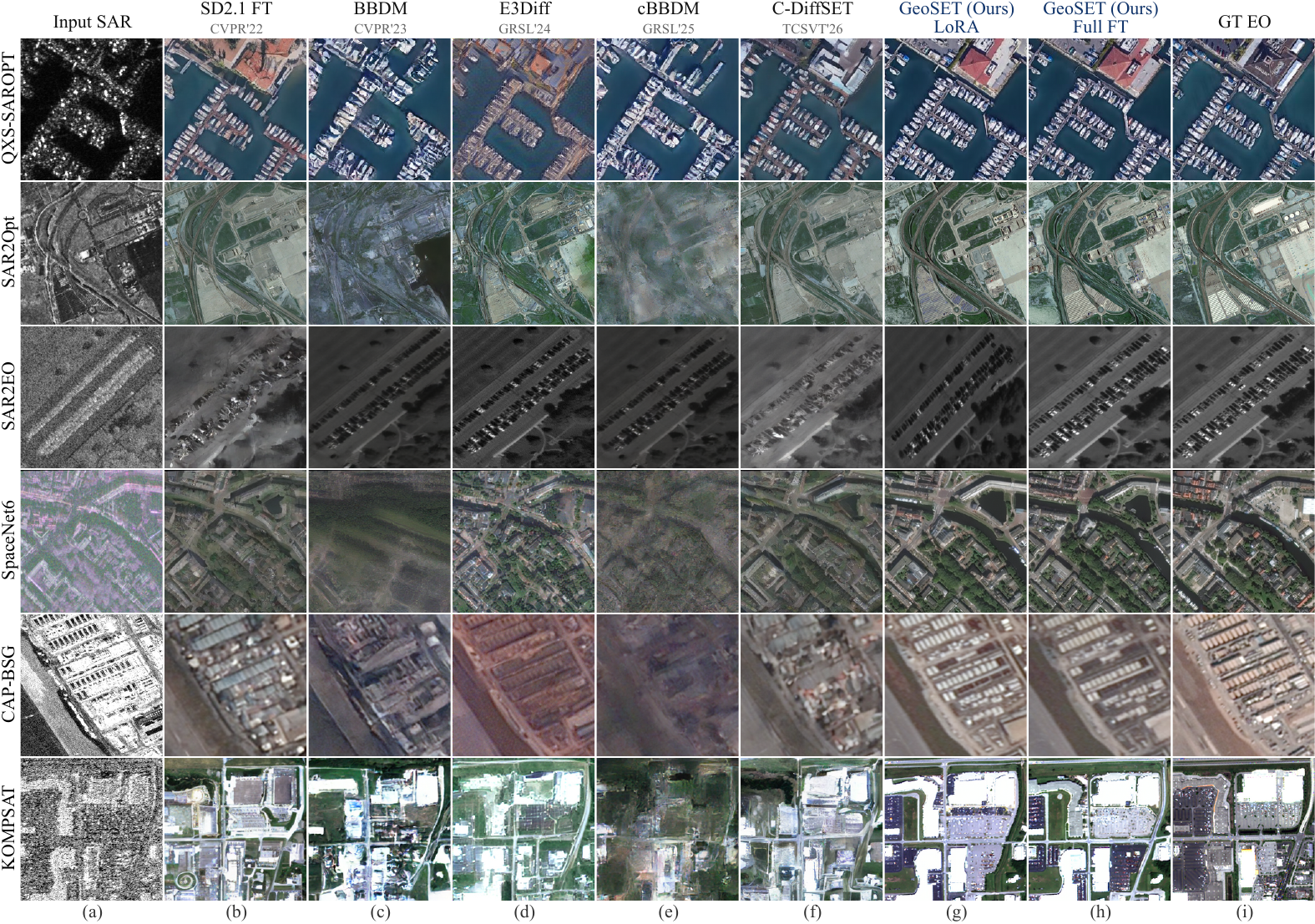}
    \vspace{-0.3cm}
    \captionof{figure}{\textbf{Qualitative comparison on six SAR-to-EO image translation (SET) benchmarks.} From left to right: (a) input SAR, (b) SD2.1 fine-tuning, (c) BBDM, (d) E3Diff, (e) cBBDM, (f) C-DiffSET, (g) GeoSET with LoRA, (h) GeoSET with full fine-tuning, and (i) ground-truth EO. GeoSET consistently produces coherent EO images while preserving fine-scale scene structure.}
    \label{fig:first}
\end{center}

\begin{abstract}
Paired synthetic aperture radar (SAR) and electro-optical (EO) imagery is increasingly available across sensors, resolutions, and geographic regions. Yet existing SAR-to-EO image translation (SET) methods are typically trained on a single, limited-scale dataset, producing models specialized to particular sensing conditions. We introduce \textbf{GeoSET}, \textit{the first generalist model for SET}, built around a single pretrained parent that is adapted to downstream datasets under a common protocol. We curate over 3 million high-quality SAR--EO pairs from a collection of more than 10 million SAR observations, spanning diverse sensors, spatial resolutions, and ground sampling distances. To bridge the modality gap between SAR observations and a pretrained image generator, we develop a speckle-robust SAR encoder and pretrain the conditional generator on this heterogeneous corpus. The resulting parent supports efficient adaptation across downstream datasets through low-rank adaptation (LoRA), updating only 0.60\% of the generator parameters and requiring approximately one hour per dataset. Across six downstream benchmarks, GeoSET achieves state-of-the-art results in FID and DISTS with full fine-tuning or LoRA, demonstrating effective transfer across heterogeneous SAR--EO domains.
\end{abstract}    
\section{Introduction}
\label{sec:introduction}

\begin{wrapfigure}{r}{0.53\columnwidth}
  \centering
  \vspace{-0.3cm}
  \includegraphics[width=0.49\columnwidth]{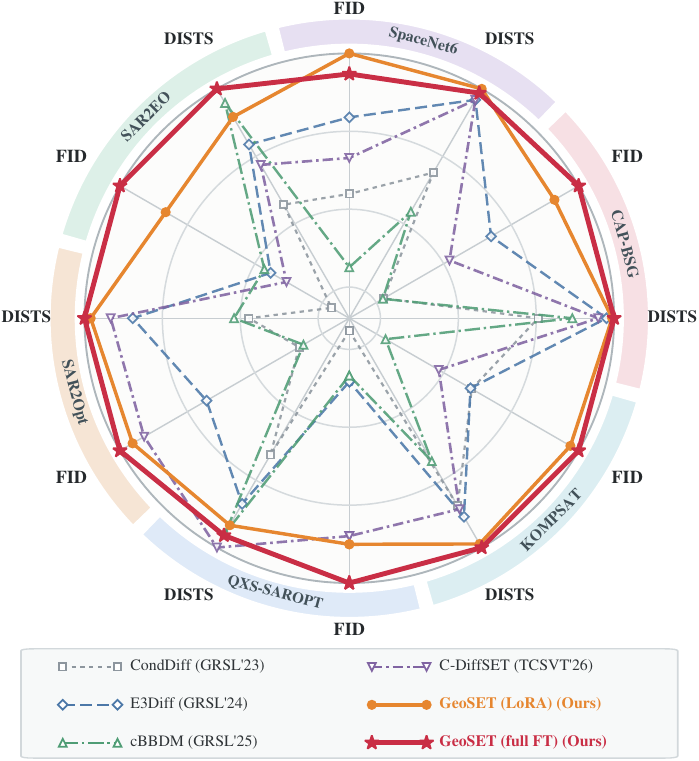}
  \caption{\textbf{Cross-dataset comparison of SAR-to-EO image translation methods.} We report FID and DISTS on six benchmarks, independently normalized for each dataset–metric pair as \(100 \times \text{best}/\text{value}\).}
  \vspace{-0.3cm}
  \label{fig:teaser}
\end{wrapfigure}

Synthetic aperture radar (SAR) enables Earth observation regardless of solar illumination and through most cloud cover~\citep{moreira2013tutorial}, making it particularly valuable when electro-optical (EO) observations are unavailable or unreliable. 
However, speckle and geometry-dependent scattering responses make SAR imagery difficult for humans to interpret and distinguish its appearance from the natural images used to pretrain modern vision models. SAR-to-EO image translation (SET) bridges this gap by mapping SAR observations into intuitive EO-like representations~\citep{zhao2022comparative,shermeyer2020spacenet6}. Such representations facilitate rapid visual analysis and may provide a familiar interface through which large-scale vision models pretrained on natural images can be applied to SAR observations~\citep{liu2024grounding,carion2026sam}. Because SAR and EO capture different physical signals, SET outputs should be interpreted as plausible SAR-conditioned visualizations rather than exact reconstructions of unobserved EO measurements.

SET has evolved from adversarial translation~\citep{turnes2020atrous,guo2024scene,lee2023segmentation} to diffusion-based generation~\citep{bai2023conditional,qin2024efficient,kim2025conditional,do2026c}. Despite these advances, existing methods typically develop a separate dataset-specific model using only the SAR--EO training pairs available in each target benchmark. Even when initialized from pretrained image generators, these models do not share a reusable SET prior learned across heterogeneous paired sources. This raises a fundamental question: \emph{Can SET learn a reusable generative prior from heterogeneous SAR--EO sources and adapt it across sensors and datasets under a common protocol?}

We address this question with \textbf{GeoSET}, to the best of our knowledge, the first generalist model for SET under a \emph{pretrain-once, adapt-many} framework. We construct a heterogeneous pretraining corpus to exploit scene structures shared across sensors, polarizations, spatial resolutions, ground sampling distances (GSDs), and geographic regions. Naively pooling these sources, however, also aggregates misregistered, low-quality, and invalid pairs. Starting from a collection of over ten million SAR observations, we identify eligible SAR--EO pairs and apply source-aware filtering to obtain over three million high-quality pairs. GeoSET is pretrained once on this corpus and subsequently adapted to every downstream benchmark using a fixed protocol for each adaptation mode. Each downstream model thus inherits a shared cross-source prior rather than learning the SAR-to-EO mapping solely from its target dataset.

Heterogeneous pretraining alone does not resolve the modality mismatch between SAR observations and generators pretrained on natural imagery. Existing latent SET methods typically use a natural-image-pretrained autoencoder to encode both the EO target and the SAR condition~\citep{kim2025conditional,do2026c,rombach2022high}. Although well suited to natural images, its encoder has not been optimized for multiplicative speckle or sensor-dependent SAR statistics. To obtain stable and informative SAR conditioning, we propose a \emph{speckle-robust SAR encoder}. The encoder learns to recover the original SAR observation from a speckle-perturbed input, encouraging it to preserve scene structures while reducing sensitivity to speckle variations. Meanwhile, keeping the pretrained decoder fixed encourages the learned SAR representation to remain compatible with its latent space, providing a common interface for SAR conditioning and EO generation.

Building on this SAR-compatible representation, GeoSET repurposes a pretrained text-to-image generator~\citep{flux-2-2025} for spatial SAR conditioning. We replace its text-conditioning stream with a SAR stream initialized from the corresponding pretrained image-stream parameters, allowing both SAR and EO inputs to be processed as spatial latent tokens. With the SAR encoder and EO autoencoder frozen, we pretrain the resulting conditional generator on the heterogeneous corpus using flow matching~\citep{lipman2022flow}. The pretrained parent supports both full fine-tuning and parameter-efficient LoRA~\citep{hu2021lora}, with LoRA adaptation requiring approximately one hour per downstream dataset.

We evaluate GeoSET on six downstream benchmarks spanning diverse sensors, resolutions, GSDs, and geographic settings. Figure~\ref{fig:teaser} summarizes its performance relative to existing SET methods. Across full fine-tuning and LoRA, GeoSET achieves the best reported FID on all six benchmarks and the best DISTS on five. LoRA updates only 0.6\% of the generator parameters while remaining competitive with full fine-tuning. Component comparisons further support the value of SAR-specific encoding and speckle augmentation. Together, these results demonstrate the effectiveness of a reusable SET prior across heterogeneous SAR--EO domains.

Our contributions are summarized as follows:
\begin{itemize}[leftmargin=1.8em]
\item We introduce \textbf{GeoSET}, to the best of our knowledge, \textit{the first generalist SET model under a pretrain-once, adapt-many framework}, pretrained on over three million SAR--EO pairs curated from heterogeneous sources through source-aware filtering.

\item We propose a two-stage pretraining strategy that combines decoder-compatible, speckle-robust SAR representation learning with multi-source pretraining of a SAR-conditioned flow model.

\item We conduct extensive evaluations and component analyses across six downstream benchmarks. GeoSET achieves state-of-the-art (SOTA) performance on most datasets, while its LoRA variant updates only 0.6\% of the generator parameters and remains competitive with full fine-tuning.
\end{itemize}
\section{Related Work}
\label{sec:related_work}

\paragraph{Conditional image generation.}
Conditional GANs established a widely used framework for paired image-to-image translation~\citep{isola2017image}, while cycle consistency enabled translation between unpaired domains~\citep{zhu2017unpaired}. Subsequent methods improved high-resolution synthesis and spatial controllability through coarse-to-fine generation and spatially adaptive normalization~\citep{wang2018high,park2019semantic}. More recently, diffusion-based approaches have advanced conditional generation through iterative denoising~\citep{saharia2022image}, Brownian-bridge formulations~\citep{li2023bbdm}, and explicit spatial conditioning of pretrained text-to-image models~\citep{zhang2023adding}. Latent diffusion reduces computational costs by operating in a compressed latent space~\citep{rombach2022high}, diffusion transformers provide scalable generative architectures~\citep{peebles2023scalable}, and flow matching offers a framework for learning continuous generative dynamics~\citep{lipman2022flow}.

\paragraph{SAR-to-EO image translation.}
Early SET methods predominantly relied on adversarial learning, incorporating multiscale context, scene-level embeddings, or EO-derived semantic supervision~\citep{turnes2020atrous,guo2024scene,lee2023segmentation}. Recent approaches adopt diffusion and bridge formulations to improve generation quality and efficiency~\citep{bai2023conditional,qin2024efficient,kim2025conditional,do2026c}. Among these, latent approaches reduce computational costs using autoencoders pretrained on natural imagery, often employing the same encoder for both the SAR condition and the EO target~\citep{kim2025conditional,do2026c}. 
However, without SAR-specific adaptation, these pretrained encoders may not adequately capture the distinct statistics of coherent radar imagery.
More importantly, existing SET methods typically optimize a separate model for each target dataset, without jointly leveraging heterogeneous SAR--EO sources to learn a shared generative prior. GeoSET addresses these limitations through speckle-robust SAR representation learning and multi-source pretraining of a reusable conditional generator.

\paragraph{Large-scale SAR--EO corpora.}
The growing availability of paired SAR--EO imagery provides new opportunities for learning cross-modal mappings across diverse sensing conditions. SAR-1M~\citep{liu2026sarmae}, although introduced for masked SAR pretraining, includes EO counterparts for a substantial portion of its SAR observations. Other recent datasets, including GUSO~\citep{yan2026guso}, TerraMesh~\citep{blumenstiel2025terramesh}, SARLO-80~\citep{debuysere2026sarlo80}, and 3MOS~\citep{ye2025threemos}, further expand paired coverage across geographic regions, sensors, spatial resolutions, and ground sampling distances. Their heterogeneity, however, makes direct pooling nontrivial: sources differ in spatial overlap, registration quality, image validity, crop density, and dataset size. GeoSET consolidates these complementary sources into a unified corpus of high-quality pairs for generative pretraining, using explicit pair filtering and source sampling based on crop-equivalent counts to account for differences in data quality, image size, and sampling density.
\section{GeoSET}
\label{sec:method}

\begin{figure*}[tbp]
  \centering
  \includegraphics[width=1.0\textwidth]{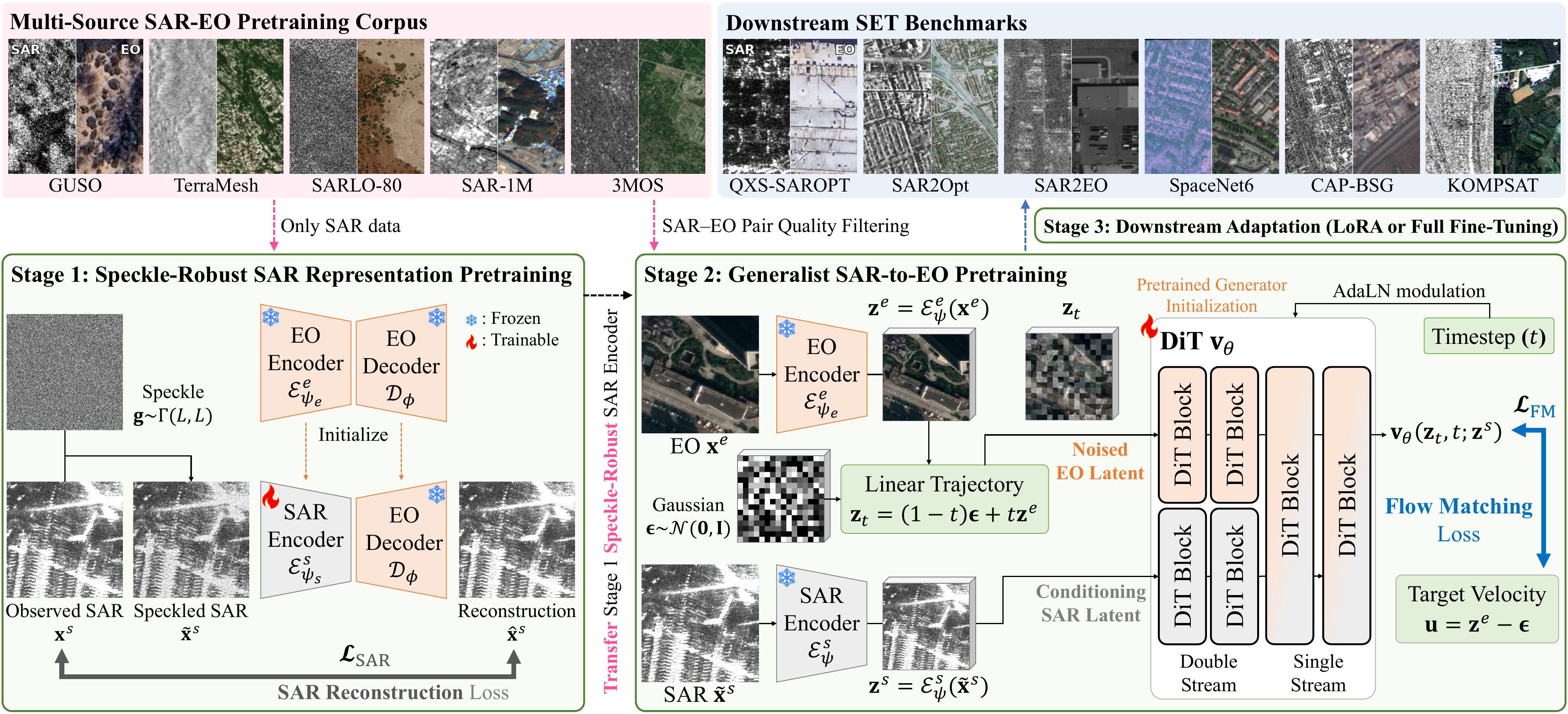}
  \caption{\textbf{GeoSET framework.} Stage~1 initializes a SAR encoder from a
  pretrained autoencoder and trains it to reconstruct the original SAR
  observation from a domain-aware speckle perturbation while the shared decoder
  remains frozen. Stage~2 replaces the pretrained generator's text condition
  with a spatial SAR stream initialized from its image stream and learns a
  conditional flow from Gaussian noise to the EO latent. Stage~3 adapts the
  resulting generalist to each downstream benchmark with LoRA or full fine-tuning.}
  \label{fig:framework}
\end{figure*}

GeoSET implements the \emph{pretrain-once, adapt-many} framework through two pretraining stages followed by downstream adaptation. Figure~\ref{fig:framework} illustrates the GeoSET framework. First, we initialize a SAR encoder from a pretrained image autoencoder and train it to reconstruct the original SAR observations from speckle-perturbed input. Reconstruction from perturbed inputs encourages speckle robustness, while keeping the decoder fixed encourages compatibility with its pretrained latent interface. Second, we convert a pretrained text-to-image generator~\citep{flux-2-2025} into a SAR-conditioned model by replacing its text-conditioning stream with a spatial SAR-conditioning stream initialized from the pretrained image stream. With the SAR encoder and EO autoencoder frozen, the resulting transformer
$\mathbf{v}_\theta$ is trained on the filtered multi-source SAR--EO corpus
using conditional flow matching~\citep{lipman2022flow}. Finally, the same generalist checkpoint is adapted to each downstream benchmark through either full fine-tuning or parameter-efficient LoRA~\citep{hu2021lora}.

\subsection{Stage 1: Speckle-Robust SAR Encoder}
\label{sec:sar_encoder}

\paragraph{Speckle-denoising reconstruction.}
Let $\mathcal{E}^{e}_{\psi_e}$ and $\mathcal{D}_{\phi}$ denote the encoder and
decoder of a pretrained image autoencoder. We initialize the SAR
encoder $\mathcal{E}^{s}_{\psi_s}$ from $\mathcal{E}^{e}_{\psi_e}$ and keep
$\mathcal{D}_{\phi}$ frozen.
For an observed SAR image $\mathbf{x}^{s}$ from source $r$, we generate
a perturbed input
$\widetilde{\mathbf{x}}^{s}
=\mathcal{A}_{\mathrm{spk}}(\mathbf{x}^{s};r)$,
where $\mathcal{A}_{\mathrm{spk}}$ is the domain-aware speckle
augmentation operator defined below.
The reconstruction is
\begin{equation}
    \widehat{\mathbf{x}}^{s}
    =
    \mathcal{D}_{\phi}
    \left(
    \mathcal{E}^{s}_{\psi_s}
    (\widetilde{\mathbf{x}}^{s})
    \right),
    \qquad
    \mathcal{L}_{\mathrm{SAR}}
    =
    \mathbb{E}_{\mathbf{x}^{s}}
    \left[
    \left\|
    \widehat{\mathbf{x}}^{s}-\mathbf{x}^{s}
    \right\|_1
    \right],
    \label{eq:sar_reconstruction}
\end{equation}
where only the SAR encoder $\mathcal{E}^{s}_{\psi_s}$ is optimized. Fixing the
decoder $\mathcal{D}_\phi$ prevents $\mathcal{E}^{s}_{\psi_s}$ and $\mathcal{D}_\phi$ from jointly forming an unconstrained
SAR-specific codec, encouraging SAR features that remain compatible with the
pretrained latent interface of $\mathcal{D}_\phi$. The reconstruction target $\mathbf{x}^{s}$ is the original SAR observation rather than a
speckle-free reference. The objective therefore encourages robustness to
the added perturbation without requiring clean SAR supervision. Stage~1 uses only SAR images
and requires neither paired EO images nor cross-modal alignment.

\paragraph{Domain-aware speckle augmentation.}
SAR sources encode radar responses in different numerical domains, making a
single multiplicative augmentation inappropriate for all datasets. For each
source $r$, let $d_r$ denote its representation domain---linear intensity,
linear amplitude, logarithmic measurement, display imagery, or unknown. We
then apply the corresponding routed operator $\mathcal{R}_{d_r}$. Following
the standard Gamma model of fully developed speckle~\citep{lee1994speckle},
we sample an i.i.d. multiplicative field
$\mathbf{g}\sim\Gamma(L,L)$ using the shape--rate parameterization,
where $L\in\{4,8,16\}$, so that each element has mean $1$ and variance $1/L$. The augmented SAR image $\widetilde{\mathbf{x}}^{s}$ is defined as
\begin{equation}
    \widetilde{\mathbf{x}}^{s}
    =
    \mathcal{A}_{\mathrm{spk}}(\mathbf{x}^{s};r)
    =
    \mathcal{R}_{d_r}(\mathbf{x}^{s};\mathbf{g}).
    \label{eq:domain_speckle}
\end{equation}
Note that this speckle augmentation operation provides a controlled robustness perturbation rather than
simulating a new lower-look acquisition, since the observed input already
contains acquisition-dependent speckle. The same augmentation is applied to SAR conditions during generalist pretraining and downstream adaptation, and is disabled at inference; detailed
routing rules are provided in the \textit{Appendix}.

\subsection{Stage 2: Generalist SAR-to-EO Pretraining}
\label{sec:generalist_pretraining}

\paragraph{SAR and EO latents.}
For each retained SAR--EO pair
$(\mathbf{x}^{s}, \mathbf{x}^{e})$, we apply the domain-aware
speckle augmentation only to the SAR observation $\mathbf{x}^{s}$, yielding
$\widetilde{\mathbf{x}}^{s}$, while leaving the paired EO target $\mathbf{x}^{e}$
unchanged. The corresponding latent representations are
\begin{equation}
    \mathbf{z}^{s}
    =
    \mathcal{E}^{s}_{\psi_s}
    \bigl(\widetilde{\mathbf{x}}^{s}\bigr),
    \qquad
    \mathbf{z}^{e}
    =
    \mathcal{E}^{e}_{\psi_e}
    \bigl(\mathbf{x}^{e}\bigr),
    \label{eq:geoset_latents}
\end{equation}
where $\mathcal{E}^{s}_{\psi_s}$ is the speckle-robust SAR encoder
learned in Stage~1 and $\mathcal{E}^{e}_{\psi_e}$ is the
pretrained EO encoder. Both encoders ($\mathcal{E}^{s}_{\psi_s},\mathcal{E}^{e}_{\psi_e}$) and decoder
($\mathcal{D}_{\phi}$) remain frozen in Stage~2.

\paragraph{Repurposing a pretrained generator for SAR conditioning.}
We repurpose a pretrained text-to-image flow transformer comprising
double-stream and single-stream blocks. The text encoder is removed,
and the text-conditioning stream is replaced with a spatial SAR stream,
while the image stream processes the noisy EO latent $\mathbf{z}_t$.
We initialize the SAR stream by copying the corresponding pretrained
image-stream parameters and retain the remaining generator weights.
The SAR and EO tokens interact in the double-stream blocks and are
jointly processed in the single-stream blocks; only the EO tokens
are passed to the final velocity head.

\paragraph{Conditional latent flow matching.}
We learn a conditional flow from an independent Gaussian source to the
EO latent $\mathbf{z}^e$. Specifically, we sample
$\boldsymbol{\epsilon}\sim\mathcal{N}(\mathbf{0},\mathbf{I})$ and
$t\sim\mathcal{U}[0,1]$, and define the linear probability path and its
target velocity as
\begin{equation}
    \mathbf{z}_{t}
    =
    (1-t)\boldsymbol{\epsilon}
    +
    t\mathbf{z}^{e},
    \qquad
    \mathbf{u}
    =
    \frac{\partial \mathbf{z}_{t}}{\partial t}
    =
    \mathbf{z}^{e}
    -
    \boldsymbol{\epsilon}.
    \label{eq:linear_flow_path}
\end{equation}
To enable classifier-free guidance (CFG)~\citep{ho2022classifier}, we independently replace the
SAR latent with an all-zero latent with probability
$p_{\mathrm{drop}}=0.1$ for each training sample.
We train the generator $\mathbf{v}_{\theta}$ to predict the target
velocity $\mathbf{u}$ by minimizing
\begin{equation}
    \mathcal{L}_{\mathrm{FM}}
    =
    \mathbb{E}
    \left[
        \left\|
            \mathbf{v}_{\theta}
            \bigl(
                \mathbf{z}_{t},
                t;
                \mathbf{z}^{s}
            \bigr)
            -
            \mathbf{u}
        \right\|_{2}^{2}
    \right].
    \label{eq:flow_matching}
\end{equation}
Because the dropped condition is represented by an all-zero SAR latent,
the generator $\mathbf{v}_\theta$ explicitly learns both conditional and null velocity fields.
At inference, we sample
$\mathbf{z}_{0}\sim\mathcal{N}(\mathbf{0},\mathbf{I})$, combine the
conditional and null predictions using two-pass CFG, integrate the resulting velocity
field from $t=0$ to $t=1$, and decode the endpoint using the frozen
decoder $\mathcal{D}_{\phi}$.

\subsection{Stage 3: Downstream Adaptation}
\label{sec:downstream_adaptation}

For each downstream dataset, we initialize GeoSET from the final exponential moving average (EMA)
checkpoint of the Stage~2 generalist model while keeping the SAR and EO
encoders and the decoder frozen. We consider two adaptation strategies:
(i) full fine-tuning (FT) that updates all generator parameters and (ii) low-rank
adaptation (LoRA)~\citep{hu2021lora} that freezes the pretrained generator weights and introduces low-rank updates to selected linear layers as
$\mathbf{W}'=\mathbf{W}+(\alpha_{L}/r_{L})\mathbf{B}\mathbf{A}$, where $\mathbf{A}$ and $\mathbf{B}$ are trainable low-rank matrices,
$r_L$ is the adapter rank, and $\alpha_L$ is the scaling factor. The adapters
are applied to the attention and MLP projections in both the
double-stream and single-stream blocks. Both strategies use the same
flow-matching objective and SAR augmentation as in Stage~2.

\begin{figure}[t]
  \centering
  \begin{minipage}{0.43\textwidth}
    \centering
    \includegraphics[width=\textwidth]{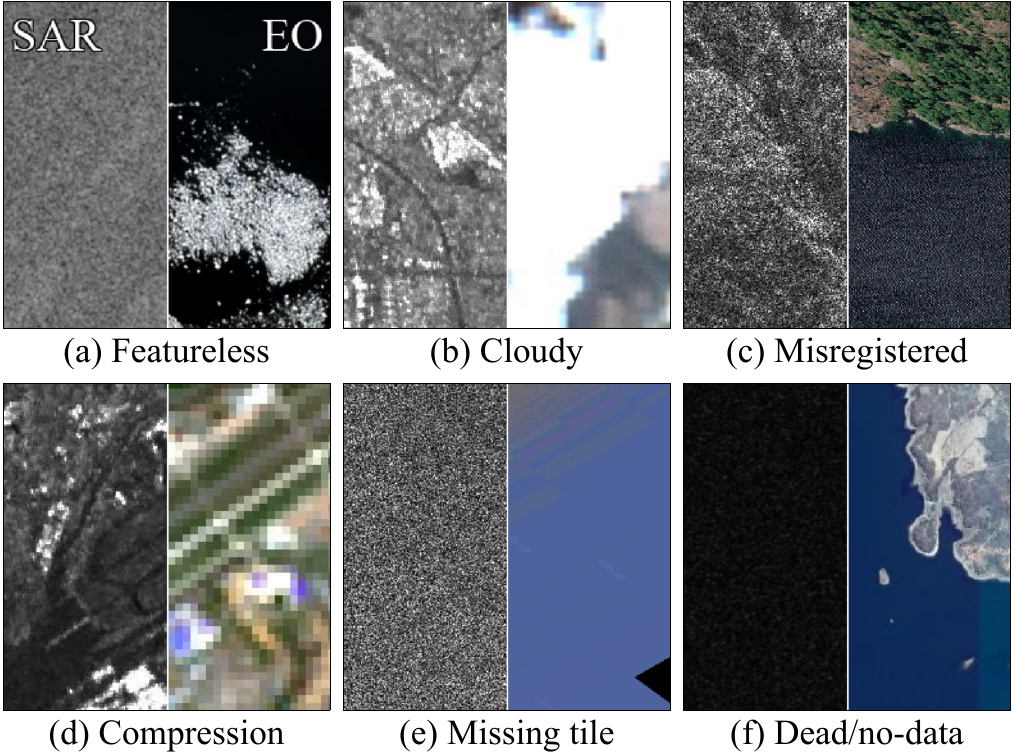}
    \caption{\textbf{Pretraining data filtering.}}
    \label{fig:filtering}
  \end{minipage}
  \hfill
  \begin{minipage}{0.54\textwidth}
    \centering
    \scriptsize
    \captionof{table}{\textbf{Stage~2 SAR--EO pretraining corpus.}
    Drop rates are computed relative to the original SAR sample counts
    and include source-specific pair eligibility selection and quality
    filtering. Crop-equivalent counts account for native image size and
    redundancy and determine the source sampling mixture.}
    \label{tab:pretrain_corpus}
    \resizebox{\linewidth}{!}{%
    \def\arraystretch{1.1}
    \setlength{\tabcolsep}{3pt}
    \begin{tabular}{lrrrrr}
      \toprule
      \textbf{Source} & \textbf{Orig. samples} & \textbf{Kept pairs} & \textbf{Drop} (\%) & \textbf{Crop equivalents} & \textbf{Mix} (\%) \\
      \midrule
      GUSO
        & 589,143 & 585,424 & 0.63
        & 2,341,696 & 38.73 \\
      TerraMesh
        & 8,194,048 & 1,741,434 & 78.75
        & 1,741,434 & 28.80 \\
      SARLO-80
        & 87,870 & 77,932 & 11.31
        & 1,246,912 & 20.62 \\
      SAR-1M
        & 1,130,379 & 688,458 & 39.09
        & 688,458 & 11.39 \\
      3MOS
        & 113,074 & 111,496 & 1.40
        & 27,874 & 0.46 \\
      \rowcolor{red!5}
      \textbf{Total}
        & 10,114,514
        & 3,204,744
        & 68.32
        & 6,046,374
        & 100.00 \\
      \bottomrule
    \end{tabular}}
  \end{minipage}
\end{figure}
\section{Experiments}
\label{sec:experiments}

\subsection{Datasets}
\label{sec:exp_data}

\paragraph{Multi-source pretraining corpus.}
We construct the pretraining corpus from GUSO~\citep{yan2026guso},
TerraMesh~\citep{blumenstiel2025terramesh},
SARLO-80~\citep{debuysere2026sarlo80},
SAR-1M~\citep{liu2026sarmae}, and 3MOS~\citep{ye2025threemos}.
Together, these sources span diverse sensors, polarizations, spatial
resolutions, ground sampling distances (GSDs), and geographic regions.
GUSO provides globally distributed, ultra-high-resolution pairs at
0.16--0.98\,m GSD; TerraMesh provides globally distributed Sentinel-1/2
observations aligned to a 10\,m grid; and SARLO-80 contains Umbra
spotlight SAR imagery resampled to a 0.8\,m slant-range grid.
SAR-1M and 3MOS further broaden the corpus with multi-sensor observations
covering different frequency bands and spatial resolutions.
Source-specific sensor, polarization, GSD, native-resolution, and
preprocessing statistics are provided in the
\textit{Appendix}.

\paragraph{Pair filtering and sampling.}
Stage~2 uses only paired SAR--EO observations. We first impose
source-specific eligibility criteria. For TerraMesh, we retain pairs with reported zero cloud cover and an acquisition-time difference of at most one day,
yielding 1,760,233 eligible pairs (21.5\%) from 8,194,048 SAR observations. For
SAR-1M, we retain the 731,073 samples (64.7\%) with an available EO counterpart from
1,130,379 SAR observations. Together with the remaining sources,
these criteria yield 3,281,393 eligible pairs. We then remove pairs exhibiting
featureless content (51,515 pairs), cloud contamination (18,722 pairs), SAR--EO
misregistration (6,413 pairs), compression artifacts (582 pairs), missing tiles (270 pairs), or
dead/no-data regions (161 pairs). It should be noted that because these categories are non-exclusive, their
counts do not sum to the number of removed pairs. Filtering the union of these non-exclusive categories leaves
3,204,744 pairs, as summarized in Table~\ref{tab:pretrain_corpus}.
Native image sizes and crop densities vary substantially across sources.
We determine source sampling probabilities from the number of
non-overlapping $256\times256$ crop equivalents.
For 3MOS, we divide the tile count by four to approximately account
for the redundancy introduced by half-tile strides along both spatial
axes. This correction reduces the overrepresentation of densely
overlapping tiles in the pretraining mixture.

\paragraph{Downstream benchmarks.}
We evaluate GeoSET on six SAR--EO benchmarks spanning satellite and airborne
platforms and GSDs from 0.5 to 1.25\,m. The train/test splits comprise
16,001/3,999 pairs for QXS-SAROPT~\citep{huang2021qxs},
1,450/627 for SAR2Opt~\citep{zhao2022comparative},
68,151/4,000 for SAR2EO~\citep{du2023sar2eo}, and
2,558/495 for SpaceNet6~\citep{shermeyer2020spacenet6}.
We additionally evaluate on two private datasets, CAP-BSG and KOMPSAT,
using train/test splits of 7,331/2,024 and 6,167/1,680 pairs, respectively.
Complete sensor, resolution, and split-construction details are provided
in the \textit{Appendix}.

\begin{table*}[tbp]
    \scriptsize
    \centering
    \caption{Quantitative comparison on QXS-SAROPT and SAR2Opt. We retrain (evaluate) all
    competing methods on the same training (test) splits. \textbf{Bold} and \underline{underlined} values indicate the best and second-best results, respectively.}
    \vspace{-0.2cm}
    \label{tab:main_standard}
    \setlength{\tabcolsep}{3.5pt}
    \renewcommand{\arraystretch}{1.1}
    \resizebox{0.95\textwidth}{!}{%
    \begin{tabular}{llcccccccccccccc}
        \toprule
        \multirow{2}{*}{\textbf{Method}}
        & \multirow{2}{*}{\textbf{Venue}}
        & \multicolumn{7}{c}{\textbf{QXS-SAROPT}}
        & \multicolumn{7}{c}{\textbf{SAR2Opt}} \\
        \cmidrule(lr){3-9}\cmidrule(lr){10-16}
        &
        & \textbf{FID}$\downarrow$
        & \textbf{DISTS}$\downarrow$
        & \textbf{KID}$\downarrow$
        & \textbf{DINO}$\uparrow$
        & \textbf{LPIPS}$\downarrow$
        & \textbf{SSIM}$\uparrow$
        & \textbf{PSNR}$\uparrow$
        & \textbf{FID}$\downarrow$
        & \textbf{DISTS}$\downarrow$
        & \textbf{KID}$\downarrow$
        & \textbf{DINO}$\uparrow$
        & \textbf{LPIPS}$\downarrow$
        & \textbf{SSIM}$\uparrow$
        & \textbf{PSNR}$\uparrow$ \\
        \midrule
        \rowcolor{black!5}
        \multicolumn{16}{l}{\textit{General image-to-image translation methods}} \\
        pix2pix & CVPR'17
        & 174.6 & 0.373 & 0.1713 & 0.275 & 0.665 & 0.203 & 12.33
        & 261.9 & 0.347 & 0.2164 & 0.277 & 0.657 & 0.199 & 13.39 \\
        CycleGAN & ICCV'17
        & 115.5 & 0.362 & 0.0802 & 0.283 & 0.647 & 0.278 & 13.25
        & 139.1 & 0.323 & 0.0343 & 0.374 & 0.642 & 0.188 & 12.68 \\
        pix2pixHD & CVPR'18
        & 85.7 & 0.298 & 0.0492 & 0.403 & 0.573 & 0.358 & 16.13
        & 146.3 & 0.283 & 0.0654 & 0.475 & 0.567 & 0.268 & 15.95 \\
        SPADE & CVPR'19
        & 90.7 & 0.292 & 0.0607 & 0.366 & 0.599 & 0.320 & 14.53
        & 142.5 & 0.265 & 0.0518 & 0.447 & 0.597 & 0.234 & 14.47 \\
        DDPM (SR3) & TPAMI'22
        & 43.8 & 0.311 & 0.0189 & 0.425 & 0.620 & 0.359 & 14.04
        & 122.5 & 0.295 & 0.0437 & 0.497 & 0.610 & 0.313 & 13.65 \\
        SD2.1 FT & CVPR'22
        & \underline{19.1} & 0.257 & \underline{0.0042} & 0.489 & 0.561 & 0.348 & 15.40
        & \underline{71.8} & 0.211 & 0.0094 & 0.600 & 0.541 & 0.293 & 16.24 \\
        BBDM & CVPR'23
        & 76.6 & 0.270 & 0.0479 & 0.414 & 0.568 & 0.352 & 15.34
        & 143.1 & 0.290 & 0.0671 & 0.466 & 0.590 & 0.276 & 15.29 \\
        ControlNet & ICCV'23
        & 50.4 & 0.307 & 0.0211 & 0.458 & 0.604 & 0.297 & 13.42
        & 140.5 & 0.350 & 0.0480 & 0.479 & 0.643 & 0.217 & 11.73 \\
        HI-Diff & NeurIPS'23
        & 324.3 & 0.539 & 0.3269 & 0.215 & 0.692 & \textbf{0.457} & \textbf{17.10}
        & 319.8 & 0.473 & 0.2357 & 0.277 & 0.692 & \textbf{0.384} & \textbf{17.36} \\
        ResShift & NeurIPS'23
        & 140.2 & 0.334 & 0.0872 & 0.295 & 0.607 & 0.217 & 14.20
        & 141.7 & 0.304 & 0.0515 & 0.435 & 0.597 & 0.177 & 14.31 \\
        StegoGAN & CVPR'24
        & 106.8 & 0.384 & 0.0707 & 0.261 & 0.658 & 0.254 & 12.96
        & 149.8 & 0.332 & 0.0396 & 0.362 & 0.652 & 0.162 & 12.39 \\
        \rowcolor{black!5}
        \multicolumn{16}{l}{\textit{SAR-to-EO image translation (SET) methods}} \\
        CondDiff & GRSL'23
        & 88.6 & 0.355 & 0.0537 & 0.310 & 0.730 & 0.213 & 11.55
        & 211.8 & 0.415 & 0.1379 & 0.343 & 0.686 & 0.248 & 12.48 \\
        E3Diff & GRSL'24
        & 47.8 & 0.278 & 0.0167 & 0.379 & \underline{0.530} & 0.302 & 16.44
        & 104.7 & 0.232 & 0.0306 & 0.541 & \textbf{0.529} & 0.249 & 16.09 \\
        cBBDM & GRSL'25
        & 50.6 & 0.246 & 0.0284 & 0.492 & 0.539 & 0.372 & 16.02
        & 222.3 & 0.377 & 0.1521 & 0.413 & 0.571 & \underline{0.361} & \underline{17.05} \\
        C-DiffSET & TCSVT'26
        & 19.9 & \textbf{0.233} & 0.0055 & 0.522 & \textbf{0.526} & \underline{0.380} & \underline{16.92}
        & 78.1 & 0.214 & 0.0138 & 0.601 & \underline{0.529} & 0.314 & 16.81 \\
        \rowcolor{orange!5}
        \textbf{GeoSET (LoRA)} & --
        & 19.3 & 0.254 & 0.0050 & \underline{0.544} & 0.564 & 0.326 & 14.67
        & 74.6 & \underline{0.200} & \underline{0.0066} & \textbf{0.626} & 0.535 & 0.278 & 15.50 \\
        \rowcolor{red!5}
        \textbf{GeoSET (full FT)} & --
        & \textbf{16.9} & \underline{0.244} & \textbf{0.0032} & \textbf{0.558} & 0.553 & 0.332 & 15.07
        & \textbf{71.1} & \textbf{0.196} & \textbf{0.0052} & \underline{0.614} & 0.532 & 0.279 & 15.66 \\
        \bottomrule
    \end{tabular}}
\end{table*}

\begin{table*}[tbp]
    \scriptsize
    \centering
    \caption{Quantitative comparison on SAR2EO and SpaceNet6. We retrain (evaluate) all
    competing methods on the same training (test) splits. \textbf{Bold} is the best, and \underline{underlined} is the 2nd best.}
    \vspace{-0.2cm}
    \label{tab:main_standard_2}
    \setlength{\tabcolsep}{3.5pt}
    \renewcommand{\arraystretch}{1.1}
    \resizebox{0.95\textwidth}{!}{%
    \begin{tabular}{lcccccccccccccc}
        \toprule
        \multirow{2}{*}{\textbf{Method}}
        & \multicolumn{7}{c}{\textbf{SAR2EO}}
        & \multicolumn{7}{c}{\textbf{SpaceNet6}} \\
        \cmidrule(lr){2-8}\cmidrule(lr){9-15}
        & \textbf{FID}$\downarrow$
        & \textbf{DISTS}$\downarrow$
        & \textbf{KID}$\downarrow$
        & \textbf{DINO}$\uparrow$
        & \textbf{LPIPS}$\downarrow$
        & \textbf{SSIM}$\uparrow$
        & \textbf{PSNR}$\uparrow$
        & \textbf{FID}$\downarrow$
        & \textbf{DISTS}$\downarrow$
        & \textbf{KID}$\downarrow$
        & \textbf{DINO}$\uparrow$
        & \textbf{LPIPS}$\downarrow$
        & \textbf{SSIM}$\uparrow$
        & \textbf{PSNR}$\uparrow$ \\
        \midrule
        \rowcolor{black!5}
        \multicolumn{15}{l}{\textit{General image-to-image translation methods}} \\
        pix2pix
        & 169.9 & 0.315 & 0.1551 & 0.400 & 0.573 & 0.455 & 18.06
        & 166.7 & 0.242 & 0.1091 & 0.589 & 0.419 & 0.459 & 17.07 \\
        CycleGAN
        & 360.7 & 0.522 & 0.3903 & 0.369 & 0.683 & 0.300 & 15.17
        & 119.3 & 0.219 & 0.0436 & 0.660 & 0.373 & 0.478 & 16.88 \\
        pix2pixHD
        & 106.0 & 0.254 & 0.0638 & 0.543 & 0.439 & 0.619 & 21.68
        & 175.1 & 0.246 & 0.1136 & 0.682 & 0.361 & \underline{0.525} & 19.21 \\
        SPADE
        & 98.5 & 0.260 & 0.0646 & 0.504 & 0.501 & 0.562 & 19.81
        & 179.7 & 0.242 & 0.1228 & 0.662 & 0.385 & 0.489 & 17.77 \\
        DDPM (SR3)
        & 85.0 & 0.272 & 0.0390 & 0.529 & 0.471 & 0.479 & 13.49
        & 219.1 & 0.333 & 0.1616 & 0.419 & 0.622 & 0.111 & 12.45 \\
        SD2.1 FT
        & 67.7 & 0.270 & 0.0282 & 0.511 & 0.487 & 0.459 & 15.46
        & 124.7 & 0.196 & 0.0537 & 0.612 & 0.348 & 0.513 & 19.29 \\
        BBDM
        & 63.7 & 0.231 & 0.0245 & 0.613 & 0.418 & 0.601 & 19.08
        & 230.2 & 0.381 & 0.1647 & 0.470 & 0.468 & 0.449 & 18.03 \\
        ControlNet
        & 87.8 & 0.315 & 0.0424 & 0.509 & 0.523 & 0.418 & 12.93
        & 131.3 & 0.279 & 0.0635 & 0.615 & 0.413 & 0.448 & 14.45 \\
        HI-Diff
        & 315.5 & 0.556 & 0.3007 & 0.155 & 0.647 & \textbf{0.672} & \underline{21.68}
        & 261.2 & 0.300 & 0.2036 & 0.499 & 0.403 & \textbf{0.589} & \textbf{20.61} \\
        ResShift
        & 130.6 & 0.273 & 0.0768 & 0.418 & 0.510 & 0.504 & 19.29
        & 155.4 & 0.230 & 0.0858 & 0.540 & 0.367 & 0.421 & 18.58 \\
        StegoGAN
        & 384.7 & 0.550 & 0.4343 & 0.387 & 0.698 & 0.180 & 12.49
        & 113.1 & 0.225 & 0.0361 & 0.649 & 0.388 & 0.457 & 16.04 \\
        \midrule
        \rowcolor{black!5}
        \multicolumn{15}{l}{\textit{SAR-to-EO image translation (SET) methods}} \\
        CondDiff
        & 113.2 & 0.317 & 0.0639 & 0.461 & 0.551 & 0.334 & 11.12
        & 160.1 & 0.274 & 0.1012 & 0.514 & 0.517 & 0.198 & 14.68 \\
        E3Diff
        & 55.5 & 0.228 & 0.0148 & 0.550 & 0.446 & 0.514 & 20.58
        & 110.6 & 0.197 & 0.0440 & 0.696 & 0.361 & 0.462 & 18.32 \\
        cBBDM
        & 52.7 & \underline{0.191} & 0.0214 & 0.675 & \textbf{0.369} & \underline{0.650} & \textbf{21.73}
        & 280.4 & 0.347 & 0.2620 & 0.450 & 0.419 & 0.447 & \underline{19.72} \\
        Seg-CycleGAN
        & -- & -- & -- & -- & -- & -- & --
        & 131.4 & 0.229 & 0.0611 & 0.649 & 0.386 & 0.471 & 16.62 \\
        C-DiffSET
        & 64.0 & 0.252 & 0.0257 & 0.527 & 0.462 & 0.512 & 17.09
        & 132.3 & 0.197 & 0.0614 & 0.611 & 0.346 & 0.520 & 19.43 \\
        \rowcolor{orange!5}
        \textbf{GeoSET (LoRA)}
        & \underline{29.5} & 0.202 & \underline{0.0042} & \underline{0.691} & 0.440 & 0.547 & 19.19
        & \textbf{87.9} & \textbf{0.189} & \textbf{0.0162} & \textbf{0.733} & \textbf{0.338} & 0.513 & 17.95 \\
        \rowcolor{red!5}
        \textbf{GeoSET (full FT)}
        & \textbf{24.5} & \textbf{0.181} & \textbf{0.0032} & \textbf{0.730} & \underline{0.399} & 0.564 & 19.72
        & \underline{94.1} & \underline{0.192} & \underline{0.0183} & \underline{0.724} & \underline{0.342} & 0.515 & 17.99 \\
        \bottomrule
    \end{tabular}}
\end{table*}

\subsection{Implementation Details and Evaluation Protocol}
\label{sec:implementation}

\paragraph{Training details.}
GeoSET is trained on $256\times256$ crops. Stages~1 and ~2 use four NVIDIA B200 GPUs.
Stage~1 is trained for 50,000 updates with a global batch size of 432,
using AdamW with a learning rate of $10^{-5}$ and a 500-update linear
warmup.
Stage~2 optimizes all 3.852B generator parameters for 500,000 updates
with a global batch size of 256, using AdamW with a peak learning rate
of $10^{-4}$ and a 1,000-update linear warmup.
Conditioning dropout is applied with probability 0.1.
Stages~1 and~2 require 14.8 and 129.3 hours of training, respectively.
Each downstream GeoSET model is adapted for 20,000 updates with a
global batch size of 16.
Full fine-tuning uses a learning rate of $2\times10^{-5}$,
whereas LoRA uses $10^{-4}$ with rank $r_{L}=16$
and scaling factor $\alpha_{L}=16$.
We clip the gradient norm to 1.0 in all stages and maintain an
exponential moving average (EMA) with decay 0.9999 during
Stages~2 and~3.
At inference, we use 50 integration steps with two-pass classifier-free guidance (CFG) and a guidance weight of 2.0.

\paragraph{Baselines.}
We retrain a broad set of baselines on the same training splits, using official implementations when available, and evaluate all outputs on identical test
pairs. General image-to-image translation baselines include
pix2pix~\citep{isola2017image}, CycleGAN~\citep{zhu2017unpaired},
pix2pixHD~\citep{wang2018high}, SPADE~\citep{park2019semantic}, and
StegoGAN~\citep{wu2024stegogan}. Restoration and conditional-generation
baselines include DDPM (SR3)~\citep{saharia2022image}, SD2.1
fine-tuning (FT)~\citep{rombach2022high}, BBDM~\citep{li2023bbdm},
ControlNet~\citep{zhang2023adding}, HI-Diff~\citep{chen2023hierarchical}, and
ResShift~\citep{yue2023resshift}. SET-specific baselines comprise
CondDiff~\citep{bai2023conditional}, E3Diff~\citep{qin2024efficient},
cBBDM~\citep{kim2025conditional}, C-DiffSET~\citep{do2026c}, and
Seg-CycleGAN~\citep{zhang2025seg}. We additionally compare FLUX.2 LoRA~\citep{flux-2-2025} with the proposed configurations
in the component analysis.

\paragraph{Metrics.}
We use FID~\citep{heusel2017gans} and DISTS~\citep{ding2020dists} as our
primary measures of distributional realism and perceptual fidelity,
respectively. We additionally report KID~\citep{binkowski2018demystifying}, DINO feature similarity computed
with a frozen DINOv3-SAT ViT-L/16 ~\citep{simeoni2025dinov3}, LPIPS~\citep{zhang2018lpips},
SSIM~\citep{wang2004image}, and PSNR.

\begin{figure*}[t]
  \centering
  \includegraphics[width=0.9\textwidth]{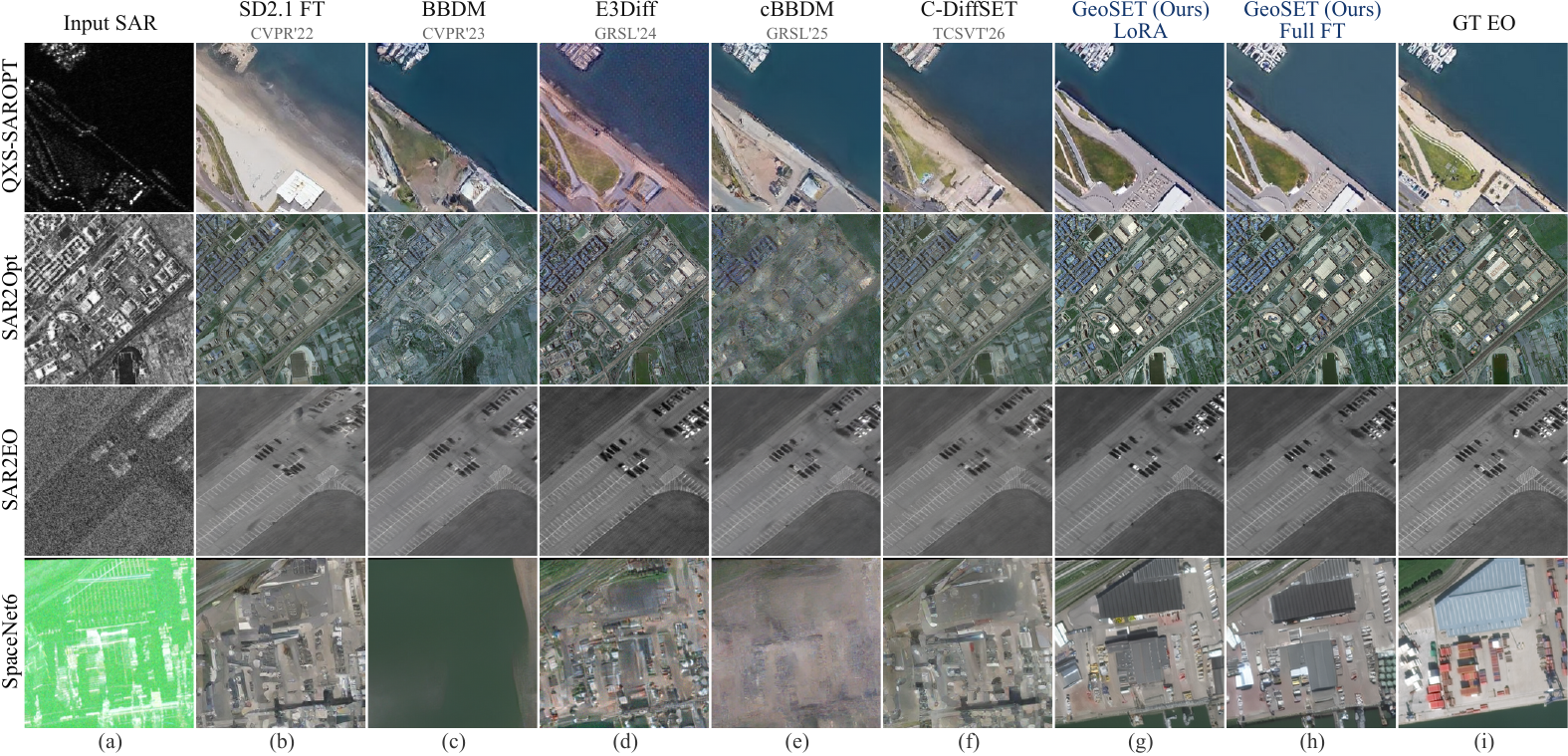}
  \caption{\textbf{Qualitative comparison on SAR-to-EO image translation (SET) benchmarks.} All methods receive the same SAR observation, and the paired EO image
  is shown as a reference.}
  \vspace{-0.2cm}
  \label{fig:result_v2}
\end{figure*}

\subsection{Comparison with the State of the Art}
\label{sec:main_results}

\paragraph{Quantitative comparison.}
Tables~\ref{tab:main_standard} and~\ref{tab:main_standard_2} compare
GeoSET with existing methods on four public benchmarks; results on
CAP-BSG and KOMPSAT are provided in the \textit{Appendix}.
Across its two adaptation modes, GeoSET achieves the best FID on all
four public benchmarks and the best DISTS on three, with gains also
observed in KID and DINO similarity.
On SAR2EO, GeoSET (full fine-tuning) reduces FID by over 50\% relative to
the strongest competing baseline, cBBDM.
On SpaceNet6, GeoSET (LoRA) achieves the strongest FID and DISTS,
demonstrating effective transfer with limited parameter updates.
C-DiffSET retains the best DISTS on QXS-SAROPT, while competing
methods often achieve stronger LPIPS, PSNR, or SSIM.
PSNR and SSIM measure agreement with a specific EO reference;
however, SAR backscatter does not uniquely determine optical color
and texture, and residual misregistration can further penalize
otherwise plausible outputs.
We therefore emphasize FID for distributional realism and DISTS
for paired structural and textural similarity~\citep{ding2020dists},
while retaining LPIPS, PSNR, and SSIM as complementary fidelity
measures. GeoSET's gains thus reflect improved distributional
quality and competitive perceptual fidelity rather than uniformly
better pixel-aligned reconstruction.

\paragraph{Qualitative comparison.}
Figures~\ref{fig:first} and ~\ref{fig:result_v2} compare fixed, score-independent examples using
identical inputs and spatial extents for all methods. We examine whether each
method preserves large-scale scene layout while producing locally coherent
building boundaries, road structures, vegetation patterns, and water regions.
This comparison complements the scalar metrics by revealing whether gains in
FID and DISTS correspond to plausible local structure rather than only changes
in global color or contrast. Additional qualitative results are provided in
the \textit{Appendix}.

\begin{table*}[t]
  \centering
  \caption{\textbf{Stage~1 encoder diagnostics and Stage~3 adaptation
  efficiency.} (a) Reconstruction PSNR on original and perturbed SAR inputs
  before and after encoder training. EO reports reconstruction with the frozen
  pretrained codec. (b) Computational costs of
  full fine-tuning and LoRA.}
  \label{tab:encoder_adaptation_summary}
  \vspace{2pt}
  \scriptsize
  \begin{minipage}[t]{0.58\textwidth}
    \vspace{-0.2cm}
    \centering
    {\small (a) Stage~1 Encoder Reconstruction}\\[2pt]
    \resizebox{\linewidth}{!}{%
    \def\arraystretch{1.1}
    \setlength{\tabcolsep}{5pt}
    \begin{tabular}{lccrccrc}
      \toprule
      \multirow{2}{*}{\textbf{Source}}
        & \multicolumn{3}{c}{\textbf{Original SAR}}
        & \multicolumn{3}{c}{\textbf{Perturbed SAR}}
        & \multirow{2}{*}{\textbf{EO}} \\
      \cmidrule(lr){2-4}\cmidrule(lr){5-7}
        & Before & After & $\Delta$
        & Before & After & $\Delta$ & \\
      \midrule
      GUSO
        & 27.47 & 29.18 & +1.71
        & 23.54 & 26.96 & +3.42
        & 31.02 \\
      TerraMesh
        & 20.11 & 32.04 & +11.93
        & 19.12 & 30.48 & +11.37
        & 33.95 \\
      SARLO-80
        & 17.39 & 19.21 & +1.82
        & 16.71 & 18.49 & +1.78
        & 27.81 \\
      SAR-1M
        & 29.95 & 31.25 & +1.30
        & 22.01 & 26.99 & +4.98
        & 33.59 \\
      3MOS
        & 35.15 & 36.07 & +0.92
        & 27.30 & 31.92 & +4.61
        & 34.62 \\
      \rowcolor{red!5}
      \textbf{Average}
        & 26.01 & 29.55 & +3.54
        & 21.74 & 26.97 & +5.23
        & 32.20 \\
      \bottomrule
    \end{tabular}}
  \end{minipage}
  \hfill
  \begin{minipage}[t]{0.36\textwidth}
    \vspace{-0.2cm}
    \centering
    {\small (b) Downstream Adaptation Efficiency}\\[2pt]
    \resizebox{\linewidth}{!}{%
    \def\arraystretch{1.1}
    \setlength{\tabcolsep}{5pt}
    \begin{tabular}{lrr}
    \toprule
    \textbf{GeoSET} & \textbf{Full FT} & \textbf{LoRA} \\
    \midrule
    Trainable parameters & 3.852B & 23.10M \\
    Trainable fraction   & 100\% & 0.5997\% \\
    Per-dataset storage  & 30.8 GB & 185 MB \\
    Peak memory          & 96.9 GB & 49.1 GB \\
    Batch size           & 16 & 16 \\
    Throughput           & 67 img/s & 94 img/s \\
    Wall time/dataset    & 1.34 h & 0.98 h \\
    \bottomrule
    \end{tabular}}
  \end{minipage}
\end{table*}

\subsection{Ablation Studies and Analysis}
\label{sec:diagnostics}

\paragraph{Stage~1 encoder reconstruction.}
Table~\ref{tab:encoder_adaptation_summary}(a) evaluates the SAR encoder
before and after Stage~1 with the decoder fixed.
Training improves reconstruction across all sources, increasing average
PSNR by 3.54\,dB on original inputs and 5.23\,dB on perturbed inputs.
The larger gain under perturbation supports speckle robustness,
while improvements on original inputs indicate better adaptation
to SAR statistics. The pretrained codec already provides strong EO
reconstruction, motivating us to preserve the EO pathway and focus
representation adaptation on SAR.

\paragraph{Parameter-efficient adaptation.}
Table~\ref{tab:encoder_adaptation_summary}(b) shows that LoRA adapts
the shared parent by updating only 0.6\% of the generator parameters,
requiring approximately one hour per dataset on one GPU.
It substantially reduces per-dataset storage and peak training
memory while increasing training throughput.
LoRA remains competitive with full fine-tuning and achieves stronger
FID, DISTS, and LPIPS on SpaceNet6 under the evaluated settings.
This makes a shared parent with compact dataset-specific adapters
a practical option for extending GeoSET to multiple downstream
datasets.

\begin{wraptable}{r}{0.61\textwidth}
    \vspace{-0.4cm}
    \scriptsize
    \centering
    \caption{\textbf{Cumulative component comparison on SAR2EO and SpaceNet6.}}
    \vspace{-0.1cm}
    \label{tab:ablation}
    \setlength{\tabcolsep}{4pt}
    \renewcommand{\arraystretch}{1.1}
    \resizebox{0.61\textwidth}{!}{%
    \begin{tabular}{lcccccc}
        \toprule
        \multirow{2}{*}{\textbf{Configuration}}
        & \multicolumn{3}{c}{\textbf{SAR2EO}}
        & \multicolumn{3}{c}{\textbf{SpaceNet6}} \\
        \cmidrule(lr){2-4}\cmidrule(lr){5-7}
        & \textbf{FID}$\downarrow$
        & \textbf{DISTS}$\downarrow$
        & \textbf{LPIPS}$\downarrow$
        & \textbf{FID}$\downarrow$
        & \textbf{DISTS}$\downarrow$
        & \textbf{LPIPS}$\downarrow$ \\
        \midrule
        SD2.1 Full FT
        & 67.7 & 0.270 & 0.487 & 124.7 & 0.196 & 0.348 \\
        \midrule
        FLUX.2 LoRA (Backbone)
        & 71.2 & 0.275 & 0.481 & 117.4 & 0.201 & 0.361 \\
        + Pretraining corpus
        & 50.5 & 0.250 & 0.470
        & 107.9 & 0.194 & 0.351 \\
        + SAR encoder
        & 39.8 & 0.220 & 0.448
        & 94.2 & 0.193 & 0.347 \\
        \rowcolor{red!5}
        + Speckle aug. (\textbf{GeoSET})
        & \textbf{29.5} & \textbf{0.202} & \textbf{0.440} & \textbf{87.9} & \textbf{0.189} & \textbf{0.338} \\
        \bottomrule
    \end{tabular}}
    \vspace{-0.2cm}
\end{wraptable}
\paragraph{Effect of pretraining, SAR encoding, and speckle augmentation.}
Table~\ref{tab:ablation} compares the cumulative effects of corpus
pretraining, SAR encoding, and speckle augmentation. All FLUX.2 configurations use matched training lengths for Stage~3
LoRA tuning and, where applicable, Stage~2 corpus pretraining
and Stage~1 SAR encoder training.
The FLUX.2 LoRA baseline shows mixed results relative to SD2.1 Full FT,
with neither configuration consistently outperforming the other.
Adding corpus pretraining improves all three metrics on both datasets,
reducing FID from 71.2 to 50.5 on SAR2EO and from 117.4 to 107.9
on SpaceNet6.
Introducing the SAR encoder further improves all metrics,
supporting the benefit of SAR-specific conditioning representations.
Speckle augmentation provides additional gains across all metrics,
yielding the best results with the complete GeoSET (LoRA) configuration.
Together, these cumulative comparisons support the complementary
benefits of corpus pretraining, SAR-specific encoding, and
speckle augmentation.
\section{Conclusion}
We introduced GeoSET, the first generalist
model for SAR-to-EO image translation under a
\emph{pretrain-once, adapt-many} framework.
By combining speckle-robust SAR encoding with generative pretraining
on over three million curated SAR--EO pairs, GeoSET learns a reusable
SET prior that transfers across heterogeneous sensing conditions.
Across six downstream benchmarks, GeoSET achieves state-of-the-art
FID and DISTS results, while LoRA enables adaptation in approximately
one hour per dataset by updating only 0.60\% of the generator parameters.
These results demonstrate the potential of a shared SET prior to
advance SAR-to-EO translation beyond dataset-specific models.

\clearpage
\appendix

\section*{Appendix}
\noindent
This \textit{Appendix} provides additional results
and implementation details for GeoSET.
Section~\ref{app:additional_results} presents quantitative and
qualitative comparisons, SAR conditioning diagnostics, and limitations.
Section~\ref{app:data_protocol} summarizes the pretraining sources
and downstream evaluation protocols.
Section~\ref{app:implementation} describes domain-aware speckle
augmentation, SAR input/output stems, and the backbone and autoencoder.
Section~\ref{app:evaluation_metrics} explains the evaluation metrics.

\begin{table}[h]
    \scriptsize
    \centering
    \caption{Overview of the \textit{Appendix}.}
    \label{tab:supple_overview}
    \resizebox{0.45\textwidth}{!}{%
    \def\arraystretch{1.1}
    \setlength{\tabcolsep}{6pt}
    \begin{tabular}{cl}
        \toprule
        \rowcolor{red!5}
        \textbf{Section} & \textbf{Contents} \\
        \midrule
        Section~\ref{app:additional_results}
        & Additional results and discussion \\
        Section~\ref{app:data_protocol}
        & Datasets and evaluation protocols \\
        Section~\ref{app:implementation}
        & Implementation details \\
        Section~\ref{app:evaluation_metrics}
        & Evaluation metrics \\
        \bottomrule
    \end{tabular}}
\end{table}

\section{Additional Results and Discussions}
\label{app:additional_results}

\paragraph{Quantitative comparison on private datasets.}
Table~\ref{tab:main_standard_3} reports results on CAP-BSG and KOMPSAT.
GeoSET achieves the best FID, DISTS, KID, and DINO similarity on both
datasets, extending the improvements observed on public benchmarks
to additional sensing conditions.
GeoSET (full fine-tuning) provides the strongest FID and DISTS, while GeoSET (LoRA)
remains competitive and matches its DINO similarity at the reported
precision.
Competing methods retain advantages in LPIPS, PSNR, and SSIM,
consistent with the different metric strengths discussed in the
main paper.

\begin{figure*}[t]
  \centering
  \includegraphics[width=0.94\textwidth]{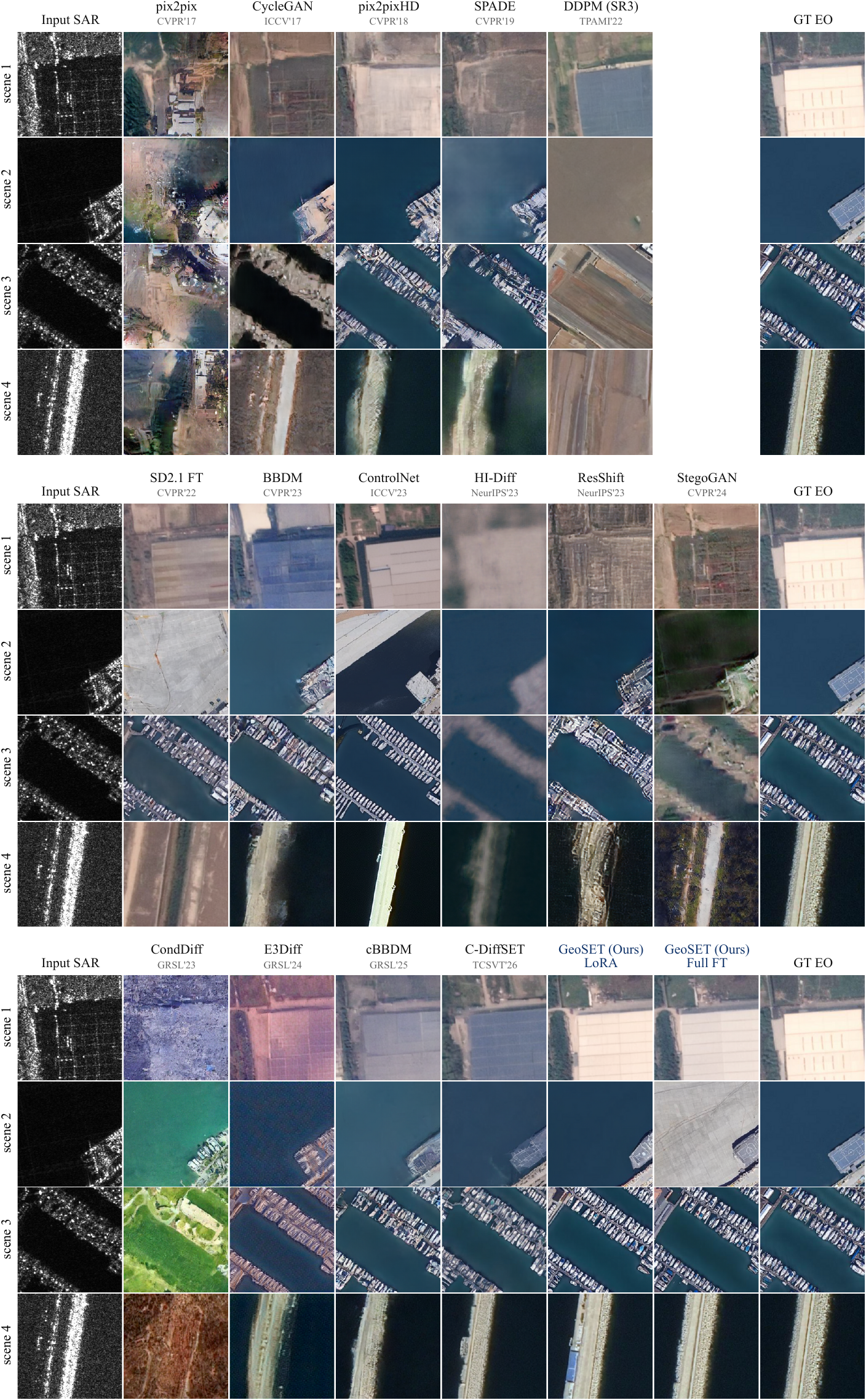}
  \caption{\textbf{Qualitative comparison on the QXS-SAROPT dataset.} All methods receive identical SAR inputs, with paired EO images shown as references.}
  \label{fig:supp_qxs}
\end{figure*}

\begin{figure*}[t]
  \centering
  \includegraphics[width=0.94\textwidth]{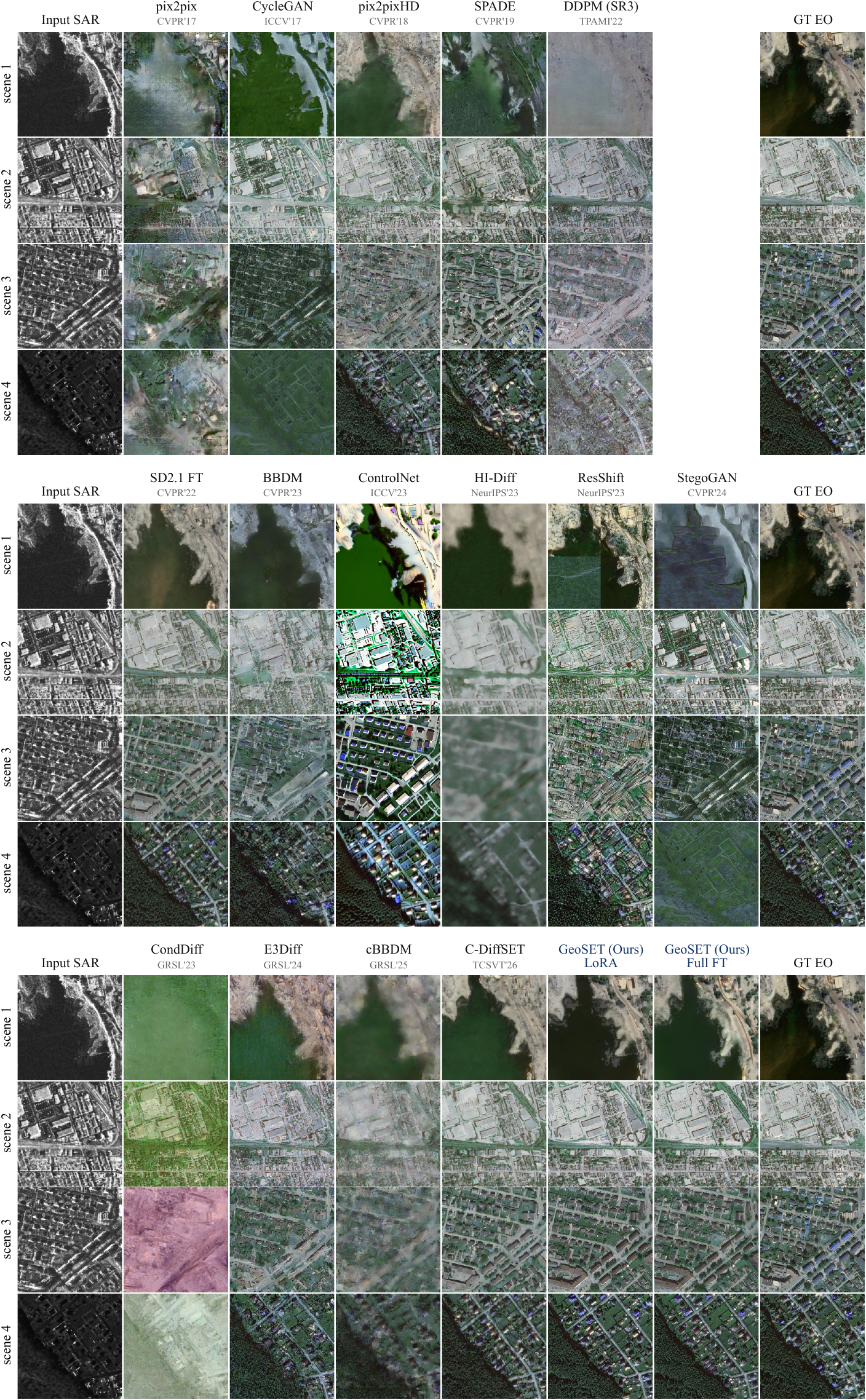}
  \caption{\textbf{Qualitative comparison on the SAR2Opt dataset.} All methods receive identical SAR inputs, with paired EO images shown as references.}
  \label{fig:supp_sar2opt}
\end{figure*}

\begin{figure*}[t]
  \centering
  \includegraphics[width=0.94\textwidth]{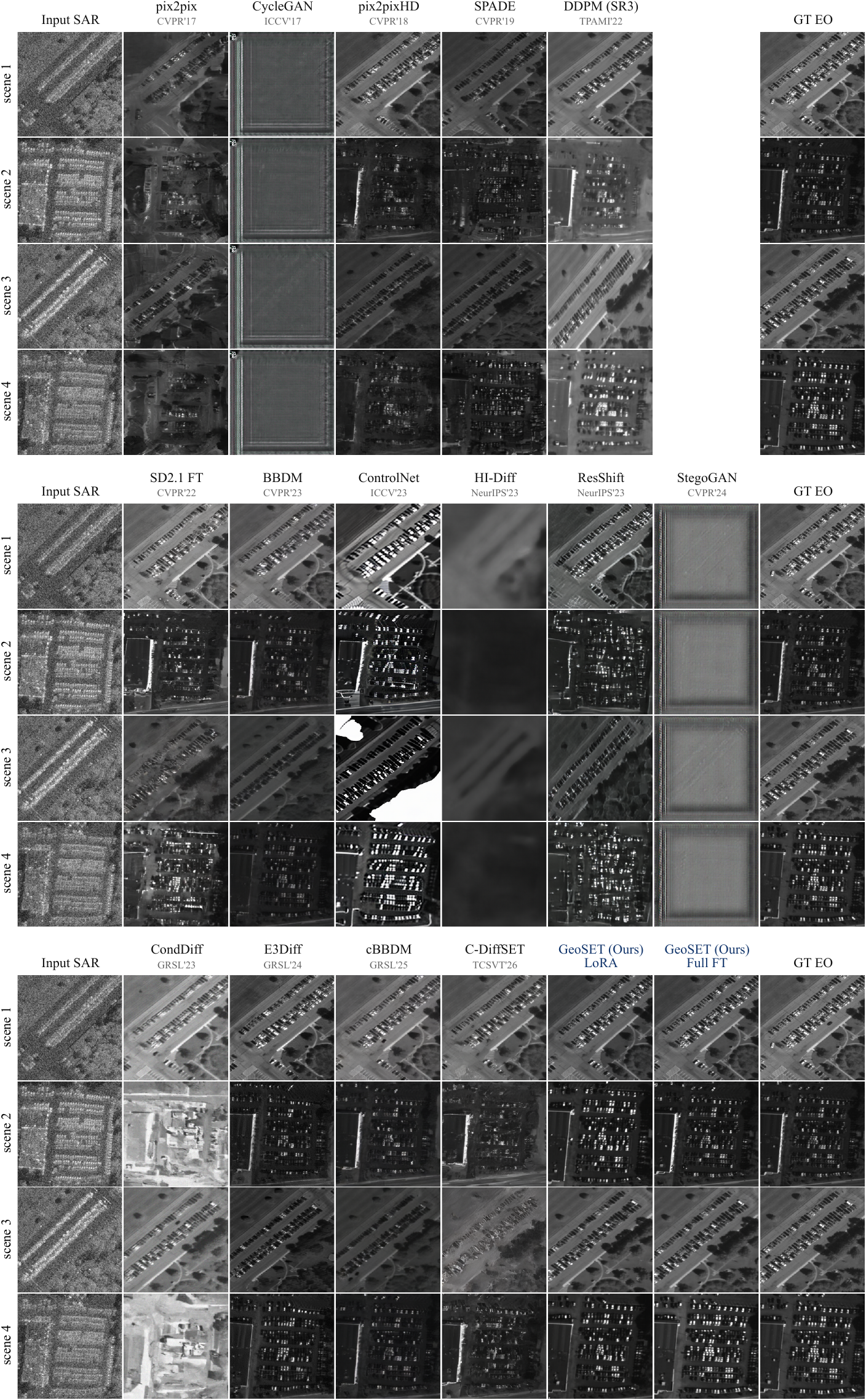}
  \caption{\textbf{Qualitative comparison on the SAR2EO dataset.} All methods receive identical SAR inputs, with paired EO images shown as references.}
  \label{fig:supp_sar2eo}
\end{figure*}

\begin{figure*}[t]
  \centering
  \includegraphics[width=0.94\textwidth]{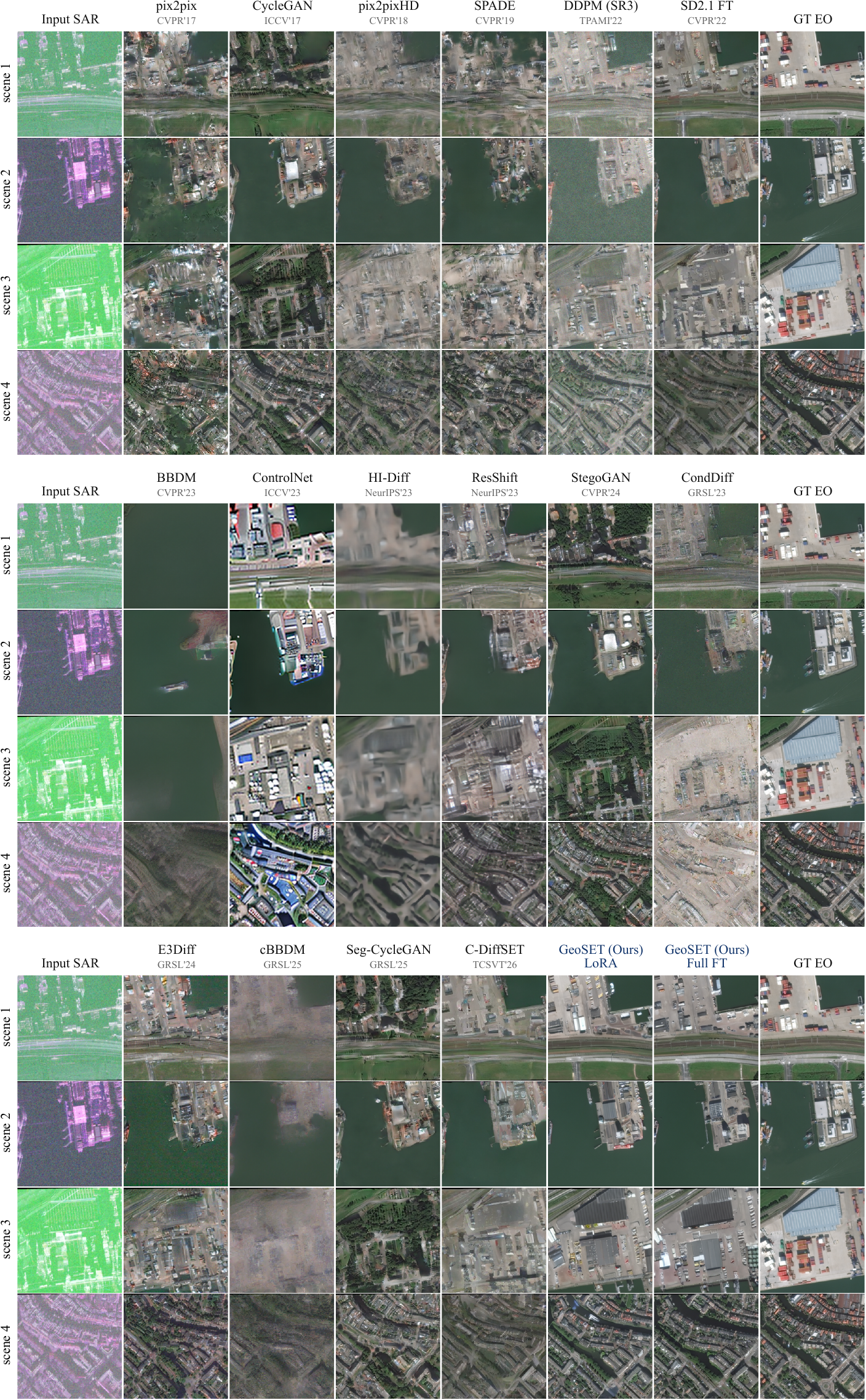}
  \caption{\textbf{Qualitative comparison on the SpaceNet6 dataset.} All methods receive identical SAR inputs, with paired EO images shown as references.}
  \label{fig:supp_spacenet}
\end{figure*}

\paragraph{Additional qualitative comparison.}
Figures~\ref{fig:supp_qxs}, ~\ref{fig:supp_sar2opt}, ~\ref{fig:supp_sar2eo}, and ~\ref{fig:supp_spacenet} present additional examples on QXS-SAROPT, SAR2Opt, SAR2EO, and SpaceNet6, respectively,
using identical SAR inputs and paired EO references.
These examples complement the quantitative results by enabling
closer inspection of scene layout, local structures, and texture
consistency across methods.

\begin{wraptable}{r}{0.57\textwidth}
    \scriptsize
    \centering
    \vspace{-0.4cm}
    \caption{\textbf{SAR conditioning intervention on pretraining sources.}
    True denotes the corresponding SAR observation, shuffled uses
    a mismatched SAR observation, and null uses an all-zero SAR latent.
    The null condition is included during training through conditioning
    dropout.}
    \label{tab:condition_intervention}
    \resizebox{0.57\textwidth}{!}{%
    \def\arraystretch{1.1}
    \setlength{\tabcolsep}{3pt}
    \begin{tabular}{lcccccc}
        \toprule
        & \multicolumn{3}{c}{\textbf{PSNR}$\uparrow$}
        & \multicolumn{3}{c}{\textbf{LPIPS}$\downarrow$} \\
        \cmidrule(lr){2-4}\cmidrule(lr){5-7}
        \textbf{Source}
        & True & Shuffled & Null
        & True & Shuffled & Null \\
        \midrule
        GUSO
        & \textbf{13.36} & 10.24 & 10.58
        & \textbf{0.548} & 0.678 & 0.690 \\
        SARLO-80
        & \textbf{15.88} & 11.96 & 11.65
        & \textbf{0.514} & 0.678 & 0.693 \\
        TerraMesh
        & \textbf{13.69} & 10.94 & 10.80
        & \textbf{0.491} & 0.670 & 0.695 \\
        \bottomrule
    \end{tabular}}
\end{wraptable}

\paragraph{Does GeoSET use the SAR condition?}
We evaluate the Stage~2 parent after 500,000 updates using true,
shuffled, and null SAR conditions.
Each setting uses the same 192 examples per source, 50 integration
steps, and random seed.
As shown in Table~\ref{tab:condition_intervention}, true conditioning
consistently improves PSNR and LPIPS over both alternatives across
all three sources.
This indicates that GeoSET uses the input SAR observation to guide
generation, rather than relying solely on its learned EO prior.
These diagnostics are conducted on pretraining sources.

\begin{figure*}[t]
  \centering
  \includegraphics[width=0.9\textwidth]{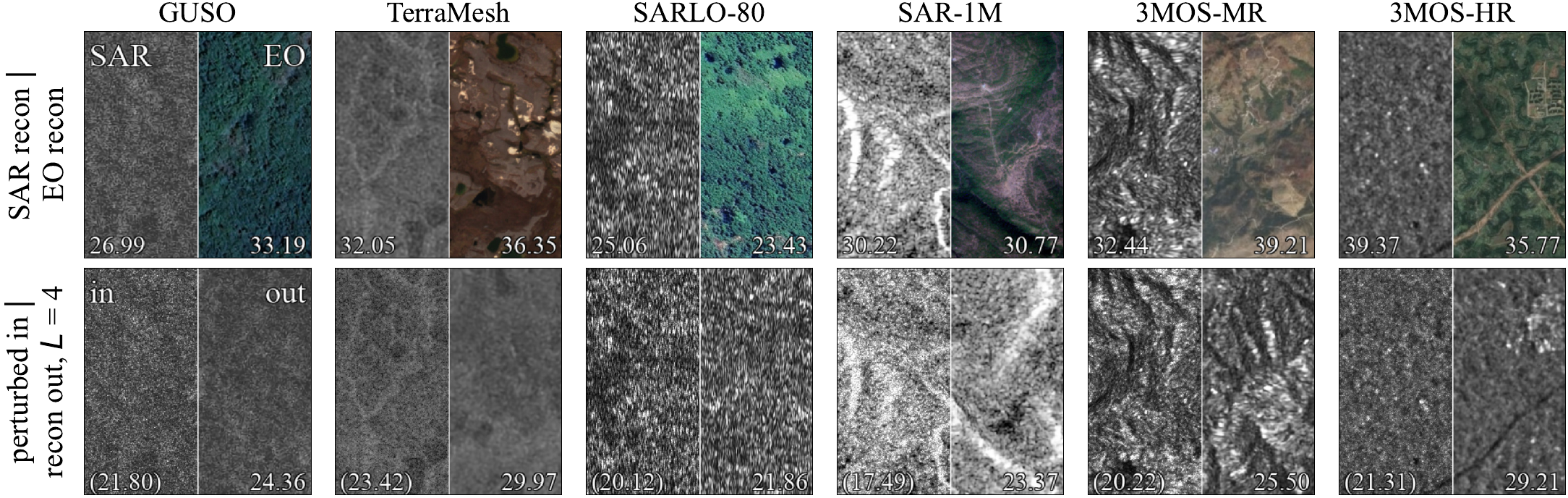}
  \caption{\textbf{SAR and EO reconstruction across pretraining sources.}
  Top: reconstructions of original SAR and EO inputs.
  Bottom: speckle-perturbed SAR inputs ($L=4$, left) and their reconstructions (right).
  Values indicate PSNR (dB) relative to the corresponding original
  observations; parenthesized values refer to perturbed inputs.}
  \label{fig:supp_encoder_output}
\end{figure*}

\paragraph{SAR and EO reconstruction.}
Figure~\ref{fig:supp_encoder_output} visualizes reconstructions
from the adapted SAR encoder and the frozen EO encoder using
their corresponding decoder paths.
Both retain scene structures across diverse pretraining sources.
For speckle-perturbed SAR inputs ($L=4$), reconstruction suppresses
the added perturbation and improves PSNR relative to the original
SAR observation in every example.
These results support robustness to additional speckle while
preserving the observed SAR structure.

\paragraph{Limitations.}
Despite heterogeneous pretraining, GeoSET does not consistently
achieve strong zero-shot performance across unseen SAR--EO domains.
Differences in sensors, acquisition geometry, resolution, and
radiometric processing remain challenges for direct transfer.
Its generalist capability therefore lies in a reusable SET prior
that enables efficient downstream adaptation, rather than universal
zero-shot translation.
LoRA makes this adaptation practical, although it still requires
paired data from the target domain.

\begin{table*}[tbp]
    \scriptsize
    \centering
    \caption{Quantitative comparison on CAP-BSG and KOMPSAT. We retrain (evaluate) all
    competing methods on the same training (test) splits. \textbf{Bold} is the best, and \underline{underlined} is the 2nd best.}
    \label{tab:main_standard_3}
    \setlength{\tabcolsep}{3.5pt}
    \renewcommand{\arraystretch}{1.1}
    \resizebox{\textwidth}{!}{%
    \begin{tabular}{lcccccccccccccc}
        \toprule
        \multirow{2}{*}{\textbf{Method}} & \multicolumn{7}{c}{\textbf{CAP-BSG}} & \multicolumn{7}{c}{\textbf{KOMPSAT}} \\
        \cmidrule(lr){2-8} \cmidrule(lr){9-15}
        & \makecell[c]{\textbf{FID}$\downarrow$}
        & \makecell[c]{\textbf{DISTS}$\downarrow$}
        & \makecell[c]{\textbf{KID}$\downarrow$}
        & \makecell[c]{\textbf{DINO}$\uparrow$}
        & \makecell[c]{\textbf{LPIPS}$\downarrow$}
        & \makecell[c]{\textbf{SSIM}$\uparrow$}
        & \makecell[c]{\textbf{PSNR}$\uparrow$}
        & \makecell[c]{\textbf{FID}$\downarrow$}
        & \makecell[c]{\textbf{DISTS}$\downarrow$}
        & \makecell[c]{\textbf{KID}$\downarrow$}
        & \makecell[c]{\textbf{DINO}$\uparrow$}
        & \makecell[c]{\textbf{LPIPS}$\downarrow$}
        & \makecell[c]{\textbf{SSIM}$\uparrow$}
        & \makecell[c]{\textbf{PSNR}$\uparrow$} \\
        \midrule
        \rowcolor{black!5}
        \multicolumn{15}{l}{\textit{General image-to-image translation methods}} \\
        pix2pix
        & 245.8 & 0.402 & 0.2086 & 0.254 & 0.626 & 0.251 & 13.77
        & 255.6 & 0.346 & 0.2892 & 0.298 & 0.669 & 0.116 & 10.33 \\
        CycleGAN
        & 100.0 & 0.327 & 0.0485 & 0.365 & 0.569 & 0.353 & 16.84
        & 160.8 & 0.359 & 0.1105 & 0.299 & 0.653 & 0.139 & 10.14 \\
        pix2pixHD
        & 212.7 & 0.335 & 0.1673 & 0.301 & 0.570 & 0.367 & \underline{17.23}
        & 223.2 & 0.379 & 0.2004 & 0.305 & 0.649 & 0.128 & 11.24 \\
        SPADE
        & 187.5 & 0.354 & 0.1407 & 0.287 & 0.593 & 0.348 & 16.20
        & 226.3 & 0.377 & 0.2106 & 0.303 & 0.670 & 0.157 & 11.47 \\
        DDPM (SR3)
        & 132.0 & 0.403 & 0.0815 & 0.297 & 0.646 & 0.373 & 12.41
        & 119.8 & 0.344 & 0.0695 & 0.365 & 0.645 & 0.121 & 10.21 \\
        SD2.1 FT
        & 98.6 & 0.348 & 0.0406 & 0.321 & 0.587 & 0.380 & 15.86
        & 100.7 & 0.333 & 0.0535 & 0.359 & 0.635 & 0.150 & 11.23 \\
        BBDM
        & 145.2 & 0.346 & 0.0789 & 0.305 & 0.577 & 0.360 & 16.16
        & 154.3 & 0.343 & 0.1101 & 0.322 & 0.656 & 0.145 & 9.74 \\
        ControlNet
        & 120.5 & 0.426 & 0.0466 & 0.290 & 0.660 & 0.303 & 13.10
        & 115.5 & 0.358 & 0.0672 & 0.357 & 0.671 & 0.137 & 9.63 \\
        HI-Diff
        & 200.0 & 0.496 & 0.1328 & 0.234 & 0.697 & \textbf{0.463} & 16.75
        & 277.8 & 0.523 & 0.2664 & 0.223 & 0.756 & \textbf{0.226} & \underline{12.39} \\
        ResShift
        & 201.2 & 0.378 & 0.1282 & 0.280 & 0.611 & 0.212 & 15.11
        & 247.9 & 0.348 & 0.2193 & 0.306 & 0.652 & 0.115 & 11.12 \\
        StegoGAN
        & 97.3 & 0.319 & 0.0464 & \underline{0.367} & 0.567 & 0.345 & 16.89
        & 325.4 & 0.497 & 0.3565 & 0.222 & 0.713 & 0.092 & \textbf{12.70} \\
        \midrule
        \rowcolor{black!5}
        \multicolumn{15}{l}{\textit{SAR-to-EO image translation (SET) methods}} \\
        CondDiff
        & 212.1 & 0.418 & 0.1347 & 0.246 & 0.734 & 0.234 & 10.50
        & 115.6 & 0.349 & 0.0672 & 0.352 & 0.692 & 0.082 & 8.70 \\
        E3Diff
        & 87.2 & 0.324 & 0.0337 & 0.346 & \textbf{0.543} & 0.332 & 16.79
        & 116.5 & 0.333 & 0.0762 & 0.388 & \textbf{0.600} & 0.147 & 11.74 \\
        cBBDM
        & 214.6 & 0.365 & 0.1580 & 0.280 & 0.578 & \underline{0.437} & \textbf{17.91}
        & 245.3 & 0.434 & 0.2300 & 0.279 & 0.666 & \underline{0.189} & 11.92 \\
        C-DiffSET
        & 113.0 & 0.333 & 0.0517 & 0.310 & 0.573 & 0.382 & 17.03
        & 144.9 & 0.344 & 0.0943 & 0.315 & 0.640 & 0.144 & 11.17 \\
        \rowcolor{orange!5}
        \textbf{GeoSET (LoRA)}
        & \underline{64.7} & \underline{0.318} & \underline{0.0154} & \textbf{0.402} & 0.566 & 0.343 & 16.85
        & \underline{72.0} & \underline{0.299} & \underline{0.0294} & \underline{0.446} & 0.625 & 0.152 & 10.64 \\
        \rowcolor{red!5}
        \textbf{GeoSET (full FT)}
        & \textbf{58.9} & \textbf{0.316} & \textbf{0.0109} & \textbf{0.402} & \underline{0.565} & 0.352 & 16.60
        & \textbf{69.8} & \textbf{0.295} & \textbf{0.0292} & \textbf{0.447} & \underline{0.620} & 0.153 & 10.78 \\
        \bottomrule
    \end{tabular}}
\end{table*}

\section{Datasets and Evaluation Protocols}
\label{app:data_protocol}

Table~\ref{tab:source_provenance} summarizes the pretraining sources,
their sensor provenance, and the SAR representations used by GeoSET.
Table~\ref{tab:downstream_splits} details the downstream datasets,
training and evaluation set sizes, and split protocols.
All competing methods are evaluated on the same test pairs
with identical crops.

\section{Implementation Details}
\label{app:implementation}

\begin{figure*}[t]
  \centering
  \includegraphics[width=0.7\textwidth]{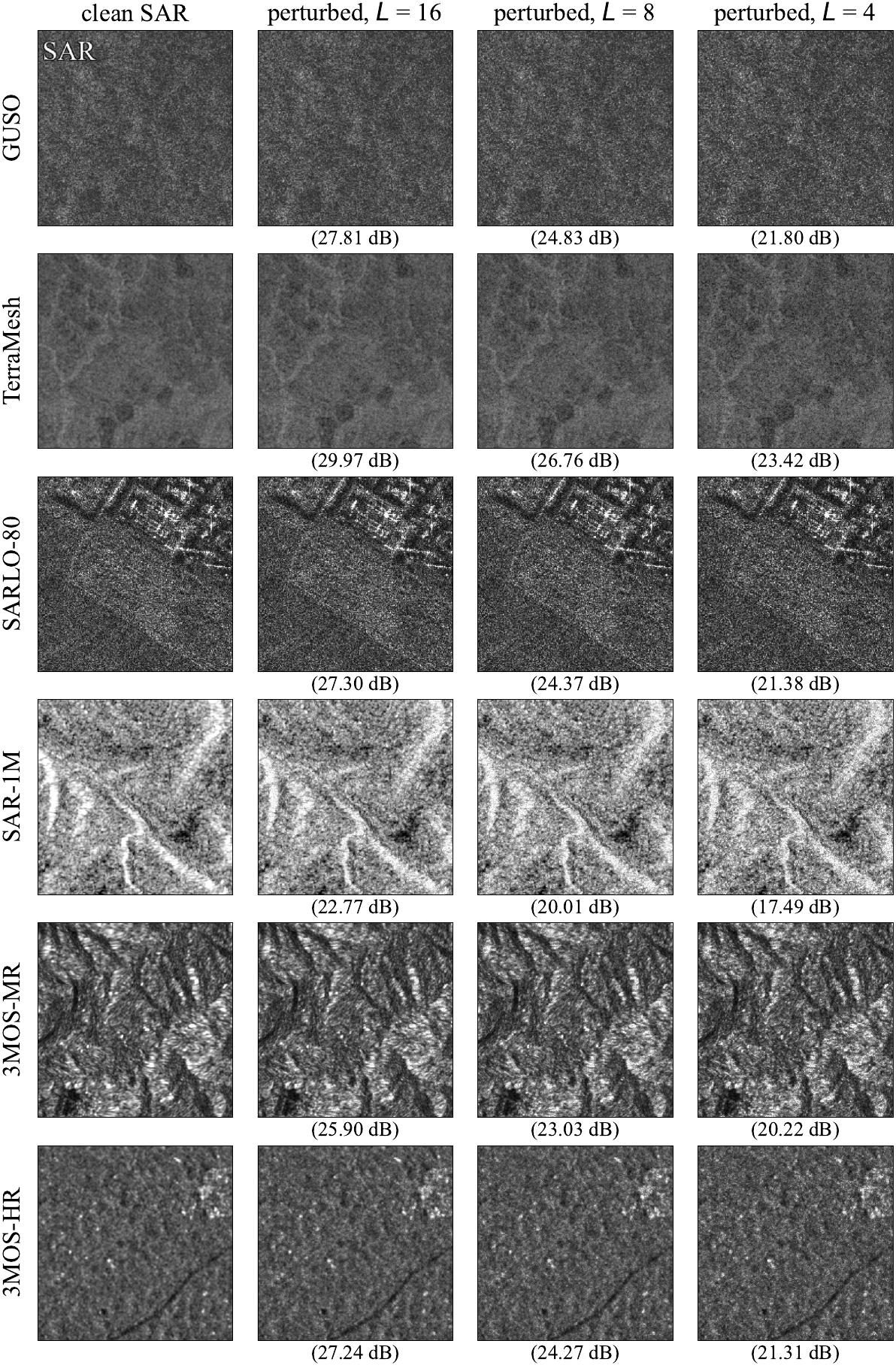}
  \caption{\textbf{Domain-aware speckle augmentation across pretraining sources.}
  From left to right: original SAR inputs and perturbed inputs with
  $L=16$, $8$, and $4$. Smaller $L$ yields stronger perturbations.
  Values indicate PSNR (dB) relative to the original SAR observations.}
  \label{fig:supp_speckle}
\end{figure*}

\paragraph{Domain-aware speckle augmentation.}
The SAR input representations summarized in
Table~\ref{tab:source_provenance} determine the routing of
$\mathcal{R}_{d_r}$ in Equation~\ref{eq:domain_speckle}.
TerraMesh uses the decibel branch, SARLO-80 uses the amplitude
branch, and GUSO, SAR-1M, and both 3MOS streams use the display branch.
The complete operator is
\begin{equation}
\mathcal{R}_{d_r}(\mathbf{x}^{s};\mathbf{g})=
\begin{cases}
  \mathbf{I}\odot\mathbf{g},
    & \text{linear intensity},\\
  \mathbf{A}\odot\sqrt{\mathbf{g}},
    & \text{linear amplitude},\\
  \mathbf{x}^{s}+10\log_{10}\mathbf{g},
    & \text{decibel-valued input},\\
  \mathbf{x}^{s}\odot[1+\kappa(\mathbf{g}-1)],
    & \text{display imagery},\\
  \mathbf{x}^{s},
    & \text{unknown representation},
\end{cases}
\label{eq:domain_speckle_v2}
\end{equation}
where $\mathbf{I}$ and $\mathbf{A}$ denote linear intensity and
amplitude, respectively, and $\odot$ denotes elementwise multiplication.
The amplitude and decibel branches follow from
$\mathbf{I}=\mathbf{A}^{2}$ and the logarithmic intensity
representation, respectively.
For display imagery, whose rendering may alter the original
radiometric relationship, we use an attenuated multiplicative
perturbation with $\kappa=0.5$.
Inputs with unknown representations are left unchanged.
Figure~\ref{fig:supp_speckle} illustrates the augmentation across
the six pretraining streams for $L\in\{16,8,4\}$.
Smaller $L$ produces stronger perturbations, reflected in lower
PSNR relative to the original SAR observations, while the
large-scale scene layout remains recognizable.

\paragraph{SAR input and output stems.}
We use channel-specific input and output stems around shared
encoder and decoder trunks to accommodate single-channel SAR
and dual-polarization VV/VH inputs.
Single-channel SAR uses the pretrained three-channel path with
grayscale folding, while VV/VH inputs use a dedicated two-channel
path that preserves separate polarization channels.
During Stage~1, we train the encoder trunk, input stems, and the
new two-channel output projection, while keeping the decoder
trunk and pretrained three-channel output projection frozen.

\paragraph{Backbone and autoencoder.}
GeoSET builds on FLUX.2 4B Base~\citep{flux-2-2025},
with five double-stream and twenty single-stream blocks,
a hidden dimension of 3,072, and 24 attention heads.
The SAR stream is initialized from the corresponding pretrained
image-stream projections, attention layers, MLPs, and modulation
parameters.
Both block types retain SAR and EO tokens and apply full
self-attention over their combined sequence, using shared spatial
rotary embeddings.
Double-stream blocks use separate projection, MLP, and modulation
parameters for the two modalities, whereas single-stream blocks
process the concatenated tokens with shared parameters.
Only EO tokens are passed to the final velocity head.
The pretrained VAE produces 32-channel posterior-mean latents
at $1/8$ spatial resolution, which are packed into 128 channels
at $1/16$ resolution and normalized using fixed BatchNorm statistics.
Decoding reverses normalization and packing.

\begin{table}[t]
    \scriptsize
    \centering
    \caption{\textbf{Pretraining corpus and representations.}
    Five dataset families form six streams, with 3MOS divided into
    \texttt{mr} (mid-resolution) and \texttt{hr} (high-resolution). Spatial scales denote reported
    resolutions or sampling intervals. Tile sizes are in pixels;
    display denotes rendered SAR imagery.}
    \label{tab:source_provenance}
    \resizebox{\textwidth}{!}{%
    \def\arraystretch{1.1}
    \setlength{\tabcolsep}{3pt}
    \begin{tabular}{llllcl}
        \toprule
        \rowcolor{red!5}
        \textbf{Source} & \textbf{SAR source} & \textbf{EO source}
        & \textbf{Spatial scale} & \textbf{Tile size} & \textbf{SAR input} \\
        \midrule
        GUSO~\citep{yan2026guso}
        & ICEYE, Capella, Umbra
        & Supplied optical imagery
        & 0.16--0.98$\mathrm{m}$
        & $512^2$
        & 1-ch display \\
        TerraMesh~\citep{blumenstiel2025terramesh}
        & Sentinel-1 RTC, VV/VH
        & Sentinel-2 L2A RGB
        & 10\,m grid
        & $264^2$
        & 2-ch dB \\
        SARLO-80~\citep{debuysere2026sarlo80}
        & Umbra spotlight SLC
        & Google satellite imagery
        & 0.8$\mathrm{m}$ slant-range grid
        & $1024^2$
        & 1-ch amplitude \\
        SAR-1M~\citep{liu2026sarmae}
        & Multi-source paired subset
        & Paired optical counterparts
        & Source-dependent
        & $256^2$
        & 1-ch display \\
        3MOS \texttt{mr}~\citep{ye2025threemos}
        & Sentinel-1, ALOS-family
        & Google Earth
        & 10 / 12.5$\mathrm{m}$
        & $256^2$
        & 1-ch display \\
        3MOS \texttt{hr}~\citep{ye2025threemos}
        & GF-3, RADARSAT-2, RCM
        & Google Earth
        & 3.5 / 6.3 / 12.5$\mathrm{m}$
        & $256^2$
        & 1-ch display \\
        \bottomrule
    \end{tabular}}
\end{table}

\begin{table}[t]
    \scriptsize
    \centering
    \caption{\textbf{Downstream datasets and evaluation protocols.}
    GSDs refer to the SAR observations where specified;
    -- denotes an unspecified value.
    SAR2EO uses the fixed first 4,000 samples of the 21,260-pair
    E3Diff test split. SAR2Opt uses a deterministic center crop.}
    \label{tab:downstream_splits}
    \resizebox{\textwidth}{!}{%
    \def\arraystretch{1.1}
    \setlength{\tabcolsep}{3pt}
    \begin{tabular}{llllcl}
        \toprule
        \rowcolor{red!5}
        \textbf{Dataset} & \textbf{SAR source} & \textbf{EO source}
        & \textbf{GSD} & \textbf{Eval. size} & \textbf{Split protocol} \\
        \midrule
        QXS-SAROPT~\citep{huang2021qxs}
        & GF-3
        & Google Earth
        & 1$\mathrm{m}$
        & $256^2$
        & C-DiffSET splits~\citep{do2026c} \\
        SAR2Opt~\citep{zhao2022comparative}
        & TerraSAR-X
        & Google Earth
        & 1$\mathrm{m}$
        & $600^2\!\to\!512^2$
        & Official train/test folders \\
        SAR2EO~\citep{du2023sar2eo}
        & Airborne UNICORN SAR
        & Monochrome WAMI
        & --
        & $256^2$
        & E3Diff split~\citep{qin2024efficient} \\
        SpaceNet6~\citep{shermeyer2020spacenet6}
        & Airborne Capella X-band
        & WorldView-2
        & 0.5$\mathrm{m}$
        & $256^2$
        & UTM-easting blocks; 450\,m guard band \\
        CAP-BSG
        & Capella X-band GEC
        & BlackSky
        & 0.9--1.25$\mathrm{m}$
        & $256^2$
        & Location-group split \\
        KOMPSAT
        & X-band SAR
        & KOMPSAT-3 RGB
        & --
        & $256^2$
        & Acquisition-group split \\
        \bottomrule
    \end{tabular}}
\end{table}

\section{Evaluation Metrics}
\label{app:evaluation_metrics}

We evaluate GeoSET using seven complementary metrics, with FID
and DISTS as the primary measures of distributional realism and
paired perceptual fidelity.
All methods are evaluated on the same test pairs and spatial extents
specified in Table~\ref{tab:downstream_splits}.
The symbols $\downarrow$ and $\uparrow$ indicate that lower and
higher values are better, respectively.

\paragraph{FID.}
Fr\'echet Inception Distance~\citep{heusel2017gans} measures the
distance between Gaussian approximations to the generated and
reference EO feature distributions.
It evaluates distributional similarity rather than correspondence
between individual SAR inputs and generated outputs.

\paragraph{DISTS.}
Deep Image Structure and Texture Similarity~\citep{ding2020dists}
compares generated and reference images through structural and
textural statistics of deep features.
Its tolerance to texture resampling complements the distributional
assessment provided by FID.

\paragraph{KID.}
Kernel Inception Distance~\citep{binkowski2018demystifying} estimates
the squared maximum mean discrepancy between generated and reference
Inception features using a polynomial kernel.
It provides a complementary distributional comparison to FID.

\paragraph{DINO similarity.}
We compare generated and paired reference EO images using features
from a frozen DINOv3-SAT ViT-L/16~\citep{simeoni2025dinov3}.
This metric assesses correspondence in a satellite-image
representation space, complementing the Inception-based
distributional measures.

\paragraph{LPIPS.}
Learned Perceptual Image Patch Similarity~\citep{zhang2018lpips}
measures differences between normalized deep features using learned
perceptual weights.
It evaluates the perceptual distance between each generated image
and its paired EO reference.

\paragraph{SSIM.}
Structural Similarity~\citep{wang2004image} compares local luminance,
contrast, and structure between generated and reference images.
It measures spatially corresponding image fidelity and is
sensitive to residual misregistration.

\paragraph{PSNR.}
Peak Signal-to-Noise Ratio expresses pixelwise mean squared error
relative to the squared image intensity range on a logarithmic scale.
It measures radiometric reconstruction fidelity to the paired EO
reference and complements the perceptual and distributional metrics.

\clearpage

\bibliography{iclr2027_conference}
\bibliographystyle{iclr2027_conference}

\end{document}